\documentclass[11pt]{article}

\usepackage{amsmath,amsbsy,amsfonts,amssymb,amsthm,dsfont,fullpage,units}
\usepackage{graphicx}
\usepackage{algorithm,algorithmic,mathtools}
\usepackage{color,cases}
\usepackage{tikz}
\usetikzlibrary{calc,shapes}
\usepackage{subfigure}
\usepackage{sublabel}
\usepackage[round]{natbib}
\usepackage{hyperref}
\usepackage{enumitem}
\usepackage{appendix}

\usepackage{array}
\newcolumntype{L}[1]{>{\raggedright\let\newline\\\arraybackslash\hspace{0pt}}m{#1}}

\usepackage{arydshln}

\newcommand{\IGNORE}[1]{}
\DeclareMathOperator{\spn}{span}

\def\nn{\nonumber}

\newcommand\E{\mathbb{E}}
\newcommand\R{\mathbb{R}}

\newcommand\poly{\operatorname{poly}}

\newcommand\eps{\varepsilon}

\newcommand\tl{\tilde}

\DeclareMathOperator{\normalize}{Norm}

\def\res{{\operatorname{res.}}}
\def\tilx{\tilde{x}}
\def\tilX{\tilde{X}}
\def\snr{{\operatorname{SNR}}}

\def\tl{\tilde}

\newcommand\inner[1]{\ensuremath{\langle #1 \rangle}}

\DeclareMathOperator{\polylog}{polylog}

\def\tha{{\mbox{\tiny th}}}

\DeclareMathOperator{\Diag}{Diag}

\def\snr{\mbox{{SNR}}}

\DeclarePairedDelimiter\norm{\lVert}{\rVert}

 \def\0{{\bf 0}}

\DeclareMathOperator{\eqd}{\overset{(d)}{=}}

\def\nn{\nonumber}

\def\qed{\hfill\hbox{${\vcenter{\vbox{
    \hrule height 0.4pt\hbox{\vrule width 0.4pt height 6pt
    \kern5pt\vrule width 0.4pt}\hrule height 0.4pt}}}$}}

\definecolor{myred}{rgb}{0.3,0.0,0.7}
\definecolor{dkg}{rgb}{0.1,0.7,0.2}
\definecolor{dkb}{rgb}{0.0,0.2,0.8}

 \def\hx{\hat{x}}

 \def\hA{\hat{A}}

\def\hT{\hat{T}}

\def\Nc{{\cal N}}

\def\Ebb{{\mathbb E}}

\def\Rbb{{\mathbb R}}

\newcommand{\bprfof}{\begin{proof_of}}
\newcommand{\eprfof}{\end{proof_of}}
\newcommand{\bprf}{\begin{myproof}}
\newcommand{\eprf}{\end{myproof}}
\newcommand{\bp}{\begin{psfrags}}
\newcommand{\ep}{\end{psfrags}}
\newcommand{\bl}{\begin{lemma}}
\newcommand{\el}{\end{lemma}}
\newcommand{\bt}{\begin{theorem}}
\newcommand{\et}{\end{theorem}}
\newcommand{\bc}{\begin{center}}
\newcommand{\ec}{\end{center}}
\newcommand{\bi}{\begin{itemize}}
\newcommand{\ei}{\end{itemize}}
\newcommand{\ben}{\begin{enumerate}}
\newcommand{\een}{\end{enumerate}}
\newcommand{\bd}{\begin{definition}}
\newcommand{\ed}{\end{definition}}
\def\beq{\begin{equation}}
\def\eeq{\end{equation}\noindent}
\def\beqn{\begin{eqnarray}}
\def\eeqn{\end{eqnarray} \noindent}
\def\beqnn{  \begin{eqnarray*}}
\def\eeqnn{\end{eqnarray*}  \noindent}
\def\bcase{  \begin{numcases}}
\def\ecase{\end{numcases}   \noindent}
\def\bsbcase{  \begin{subnumcases}}
\def\esbcase{\end{subnumcases}   \noindent}

\newtheorem{theorem}{Theorem}
\newtheorem{corollary}{Corollary}
\newtheorem{lemma}[theorem]{Lemma}

\newtheorem{claim}{Claim}

\newtheorem{definition}{Definition}

\newenvironment{myproof}{\noindent{\bf Proof:} \hspace*{1em}}{
    \hspace*{\fill} $\Box$ }
\newenvironment{proof_of}[1]{\noindent {\bf Proof of #1: }}{\hspace*{\fill} $\Box$ }

\newcommand{\matplottc}[1]{               
        \unitlength .45truein
        \begin{center}
        \includegraphics{#1.ps}
        \end{picture}
        \end{center}
}

\def\psfancypar#1#2{\begingroup\def\par{\endgraf\endgroup\lineskiplimit=0pt}
               \setbox2=\hbox{\large\sc #2}
               \newdimen\tmpht \tmpht \ht2 \advance\tmpht by \baselineskip
               \font\hhuge=Times-Bold at \tmpht
               \setbox1=\hbox{{\hhuge #1}}
               \count7=\tmpht \count8=\ht1
               \divide\count8 by 1000 \divide\count7 by \count8
               \tmpht=.001\tmpht\multiply\tmpht by \count7
               \font\hhuge=Times-Bold at \tmpht
               \setbox1=\hbox{{\hhuge #1}}
               \noindent
                \hangindent1.05\wd1
               \hangafter=-2 {\hskip-\hangindent
               \lower1\ht1\hbox{\raise1.0\ht2\copy1}%
                \kern-0\wd1}\copy2\lineskiplimit=-1000pt}

\def\Kout{\setbox1=\hbox{\Huge\bf K}\hbox to
1.05\wd1{\hspace{.05\wd1}
}}
\def\Sout{\setbox1=\hbox{\Huge\bf S}\hbox to 1.05\wd1{\hspace{.05\wd1}
}}

\newcommand{\torestate}[3]{%
\expandafter \def \csname BBRESTATE #2 \endcsname{#3}
\theoremstyle{plain}
\newtheorem{BBRESTATETHMNUM#2}[theorem]{#1}
\begin{BBRESTATETHMNUM#2}\label{#2}\csname BBRESTATE #2 \endcsname   \end{BBRESTATETHMNUM#2}
\newtheorem*{BBRESTATETHMNONNUM#2}{{#1}~\ref{#2}}
}

\newcommand{\restate}[1]{\begin{BBRESTATETHMNONNUM#1}[Restated] \csname BBRESTATE #1 \endcsname
\end{BBRESTATETHMNONNUM#1}}

\author{Anima Anandkumar\footnote{University of California, Irvine. Email: a.anandkumar@uci.edu} \and Rong Ge\footnote{Duke University. Email: rongge@cs.duke.edu} \and Majid Janzamin\footnote{University of California, Irvine. Email: mjanzami@uci.edu}}

\title{Analyzing Tensor Power Method Dynamics \\ in Overcomplete Regime}

\begin{document}
\maketitle

\begin{abstract}
We present a novel analysis of the dynamics of tensor power iterations in the overcomplete regime where the tensor CP rank is larger than the input dimension. Finding the CP decomposition of an overcomplete tensor is NP-hard in general. We consider the case where the tensor components are randomly drawn, and show that the simple power iteration recovers the components with bounded error under mild initialization conditions. We   apply  our analysis to unsupervised learning of   latent variable models,  such as   multi-view mixture models and spherical Gaussian mixtures. Given the third order moment tensor, we learn the parameters using tensor power iterations. We prove it can correctly  learn the model parameters when the number of hidden components $k$ is much larger than the data dimension $d$, up to $k = o(d^{1.5})$. We initialize the power iterations with data samples and prove its success under mild conditions on the signal-to-noise ratio of the samples. Our analysis significantly expands the class of latent variable models where spectral methods are applicable. Our analysis also deals with noise in the input tensor leading to sample complexity result in the application to learning latent variable models.
\end{abstract}


\section{Introduction}
CANDECOMP/PARAFAC (CP) decomposition of a symmetric tensor $T \in \R^{d \times  d \times d}$ is the process of decomposing it into  a succinct  sum of rank-one tensors, given by
\begin{equation} \label{eqn:tensordecomp-intro}
T = \sum_{j\in [k]} \lambda_j a_j \otimes a_j \otimes a_j, \quad \lambda_j \in \Rbb, \ a_j \in \Rbb^d,
\end{equation}
where $\otimes$ denotes the outer product. The minimum $k$ for which the tensor can be decomposed in the above form is called the (symmetric) tensor rank. 
{\em Tensor power iteration} is a simple, popular and efficient method for recovering the tensor rank-one components $a_j$'s. The tensor power iteration is given by
\begin{equation} \label{eqn:poweriteration}
x \leftarrow \frac{T(I,x,x)}{\|T(I,x,x)\|},
\end{equation}
where 
$$T(I,x,x) :=  \sum_{j,l \in [d]} x_j x_l T(:,j,l) \in \R^d$$ is a {\em multilinear} combination of tensor {\em fibers}, and $\|\cdot\|$ is the $\ell_2$ norm operator.
See Section~\ref{sec:notations} for an overview of tensor notations and preliminaries.

The tensor power iteration is a generalization of matrix power iteration: for matrix $M \in \R^{d \times d}$, the power iteration is given by $x \leftarrow Mx/\|Mx\|$. Dynamics and convergence properties of  matrix power iterations are well understood~\citep{horn2012matrix}. On the other hand, a   theoretical understanding of tensor power iterations is much more limited. Tensor power iteration can be viewed as a {\em gradient descent} step (with infinite step size), corresponding to the problem of finding the best rank-$1$ approximation of the input tensor $T$~\citep{JMLR:v15:anandkumar14b}. This optimization problem is non-convex. 
Unlike the matrix case, where the number of isolated stationary  points of power iteration is at most the dimension (given by  eigenvectors corresponding to unique eigenvalues), in the  tensor case, the number of stationary points is, in fact, exponential in the input dimension~\citep{CartwrightSturmfels2013}. This makes the analysis of tensor power iteration far more challenging.

Despite the above challenges, many advances have been made in understanding the tensor power iterations in specific regimes. When the components $a_j$'s are orthogonal to one another, it is known that there are no spurious local optima for tensor power iterations, and the only stable fixed points correspond to the true $a_j$'s~\citep{ZG01,JMLR:v15:anandkumar14b}. Any tensor with linearly independent components $a_j$'s can be orthogonalized, via an invertible transformation  (whitening) and thus, its components can be recovered efficiently. A careful perturbation analysis in this setting was carried out in~\citet{JMLR:v15:anandkumar14b}.

The framework in~\citet{JMLR:v15:anandkumar14b} is  however not applicable in the overcomplete setting, where the tensor rank $k$ exceeds the dimension $d$. Such overcomplete tensors cannot be orthogonalized and finding guaranteed decomposition is a challenging open problem. It is known that finding CP tensor decomposition is NP-hard~\citep{TensorNPHard}. In this paper, we make significant headway in showing that the simple power iterations can recover the components in the overcomplete regime under a set of mild conditions on the components $a_j$'s.

Overcomplete tensors also arise in many machine learning applications such as moments of many latent variable models, e.g., multiview mixtures, independent component Analysis (ICA), and sparse coding models, where the number of hidden variables exceeds the input dimensions~\citep{AltTensorDecomp:COLT2015}. Overcomplete models often have impressive empirical performance~\citep{coates2011analysis}, and can provide greater flexibility in modeling, and are   more robust to noise~\citep{lewicki2000learning}. By studying algorithms for overcomplete tensor decomposition, we expand the class of models that can be learnt efficiently using simple spectral methods such as tensor power iterations. Note there are other algorithms for decomposing overcomplete tensors~\citep{de2007fourth, fourierpca,bhaskara2013smoothed}, but they all require tensors of at least $4$-th order and require large computational complexity. \citet{GeMa2015} works for 3rd order tensor but requires quasi-polynomial time. The main contribution of this paper is an analysis for the practical power method in the overcomplete regime.

\subsection{Summary of results} \label{sec:summary}


We analyze the dynamics of third order tensor power iterations in the overcomplete regime. 
We assume that the tensor components $a_j$'s are randomly drawn from the unit sphere.
Since general tensor decomposition is challenging in the overcomplete regime, we argue that this is a natural first step to consider for tractable recovery.

We characterize the basin of attraction for the local optima near the rank-one components $a_j$'s. We show that under mild initialization condition, there is fast convergence to these local optima in   $O(\log \log d)$ iterations (i.e., quadratic convergence as opposed to linear convergence in case of matrices). This result is the core technical analysis of this paper stated in the following theorem. 

\begin{theorem}[Dynamics of tensor power iteration] \label{thm:dynamics}
Consider tensor $\hT = T+E$ such that exact tensor $T$ has rank-$k$ decomposition in~\eqref{eqn:tensordecomp-intro} 
with rank-one components $a_j \in \R^d, j \in [k]$ being uniformly i.i.d.\ drawn from the unit $d$-dimensional sphere, and the ratio of maximum and minimum (in absolute value) weights $\lambda_j$'s being constant. In addition, suppose the perturbation tensor $E$ has bounded norm as
\begin{equation} \label{eqn:perturbation-bound}
\|E\| \leq \epsilon \frac{\sqrt{k}}{d}, \quad \textnormal{where} \quad \epsilon < o\left( \frac{\sqrt{k}}{d} \right).
\end{equation}
Let tensor rank $k =o(d^{1.5})$, and the unit-norm initial vector $x^{(1)}$ satisfy the correlation bound
\begin{equation}\label{eqn:init}
|\inner{x^{(1)},a_j}| \geq d^\beta\frac{\sqrt{k}}{d},
\end{equation}
w.r.t.\ some true component $a_j, j \in [k]$, for some constant $\beta>0$. After $N = \Theta \left( \log \log d \right)$ iterations, the tensor power iteration  in \eqref{eqn:poweriteration} outputs a vector having w.h.p.\  a constant correlation with the true component $a_j$ as
$|\inner{x^{(N+1)},a_j}| \ge 1-\gamma$, for any fixed constant $\gamma>0$.
\end{theorem}




As a corollary, this result can be used for learning  latent variable models such as  multiview mixtures.  We show that the above initialization condition is satisfied using a sample with mild signal-to-noise ratio; see Section~\ref{sec:learning} for more details on this.

The above result is  a significant improvement over the recent analysis by~\citet{AltTensorDecomp:COLT2015,AltTensorDecomp2014,OvercompleteLVMs2014} for overcomplete tensor decomposition. In these works, it is required for the initialization vectors to have a constant amount of correlation with the true $a_j$'s. However, obtaining such strong initializations is usually not realistic in practice. On the other hand, the initialization condition in \eqref{eqn:init} is mild, and decaying even when the rank $k$ is significantly larger than dimension $d$; up to $k=o(d^{1.5})$. In learning the mixture model, such initialization vectors can be obtained as samples from the mixture model, even when there is a large amount of noise. Given this improvement, we combine our analysis in Theorem~\ref{thm:dynamics}, and the guarantees in \citep{AltTensorDecomp2014}, proving that the model parameters can be recovered consistently.

A detailed proof outline for Theorem~\ref{thm:dynamics} is provided in Section~\ref{sec:proofoutline}.
Under the random assumption, it is not hard to show that the first iteration of tensor power update makes progress. However, after the first iteration, the input vector and the tensor components are no longer {\em independent} of each other. Therefore, we cannot directly repeat the same argument for the second step. 

How do we analyze the second step even though the vector and tensor components are correlated? The main intuition is to characterize the dependency between the vector and the tensor components, and show that there is still enough randomness left for us to repeat the argument.
This idea was inspired by the analysis of Approximate Message Passing (AMP) algorithms~\citep{AMP2010}.   However, our analysis here is very different in several key aspects: 1) In approximate message passing, typically the analysis works in the {\em large system limit}, where the number of iterations is fixed and the dimension   goes to infinity. Here we can handle  a superconstant number of iterations $O(\log \log d)$, even for finite $d$; 2) Usually $k$ is assumed to be a constant factor times $d$ in the AMP-like analysis, while here we allow them to be polynomially related.

\subsection{Related work}


\paragraph{Tensor decomposition for learning latent variable models:}
In the introduction, some related works are mentioned which study the theoretical and practical aspects of spectral techniques for learning latent variable models. Among them, \citet{JMLR:v15:anandkumar14b} provide the analysis of tensor power iteration for learning several latent variable models in the undercomplete regime. \citet{AltTensorDecomp2014} provide the analysis in the overcomplete regime and \citet{OvercompleteLVMs2014} provide tensor concentration bounds and apply the analysis in~\citep{AltTensorDecomp2014} to learning LVMs proposing tight sample complexity guarantees.


\paragraph{Learning mixture of Gaussians:}Here, we provide a subset of related works studying learning mixture of Gaussians which are more comparable with our result. For a more detailed list of these works, see \citet{JMLR:v15:anandkumar14b,hsu2013learning}.
The problem of learning mixture of Gaussians dates back to the work by \citet{Pearson1895}. They propose a moment-based technique that involves solving systems of multivariate polynomials which is in general challenging in both computational and statistical sense. Recently, lots of studies on learning Gaussian mixture models have been done improving both aspects which can be divided to two main classes: distance-based and spectral methods.

Distance-based methods impose separation condition on the mean vectors showing that under enough separation the parameters can be estimated. Among such approaches, we can mention \citet{Dasgupta:GaussianMixture,VempalaWang:GaussianMixture,AroraKannan:Mixtures}. As discussed in the summary of results, these results work even if $k>d^{1.5}$ as long as the separation condition between means is satisfied, but our work can tolerate higher level of noise in the regime of $k=o(d^{1.5})$ with polynomial computational complexity. The guarantees in \citep{VempalaWang:GaussianMixture} also work in the high noise regime but need higher computational complexity as polynomial in $k^{O(k)}$ and $d$.

In the spectral approaches, the observed moments are constructed and the spectral decomposition of the observed moments are performed to recover the parameters~\citep{KalaiEtal:GaussianMixture,AnandkumarHsuKakade:COLT12,OvercompleteLVMs2014}.
\citet{KalaiEtal:GaussianMixture} analyze the problem of learning mixture of two general Gaussians and provide algorithm with high order polynomial sample and computational complexity. Note that in general, the complexity of such methods grow exponentially with the number of components without further assumptions~\citep{MoitraValiant:GaussianMixture}.
\citet{hsu2013learning} provide a spectral algorithm under non-degeneracy conditions on the mean vectors and providing guarantees with polynomial sample complexity depending on the condition number of the moment matrices. \citet{OvercompleteLVMs2014} perform tensor power iteration on the third order moment tensor to recover the mean vectors in the overcomplete regime as long as $k = o(d^{1.5})$, but  need very good initialization vector having constant correlation with the true mean vector. Here, we improve the correlation level required for convergence.

\subsection{Notation and tensor preliminaries} \label{sec:notations}
Let $[k] := \{1,2,\dotsc,k\}$, and $\|v\|$ denote the $\ell_2$ norm of vector $v$. We use $\tl{O}$ and $\tl{\Omega}$ to hide $\polylog$ factors in asymptotic notations $O$ and $\Omega$, respectively.

\paragraph{Tensor preliminaries:}
A real \emph{$p$-th order tensor} $T \in \bigotimes^p \R^d$ is a member of the outer product of Euclidean spaces $\R^{d}$.
The different dimensions of the tensor are referred to as {\em modes}. For instance, for a matrix, the first mode refers to columns and the second mode refers to rows.
In addition,￼ {\em fibers} are higher order analogues of matrix rows and columns. A fiber is obtained by fixing all but one of the indices of the tensor (and is arranged as a column vector). For example, for a third order tensor $T\in \R^{d \times d \times d}$, the mode-$1$ fiber is given by $T(:, j, l)$. 
Similarly, {\em slices} are obtained by fixing all but two of the indices of the tensor. For example, for the third order tensor $T$, the slices along $3$rd mode are given by $T(:, :, l)$.

We view a tensor $T \in \Rbb^{d \times d \times d}$ as a multilinear form. In particular, for vectors $u,v,w \in \R^d$, we have\,\footnote{Compare with the matrix case where for $M \in \R^{d \times d}$, we have $ M(I,u) = Mu := \sum_{j \in [d]} u_j M(:,j)$.}
\begin{equation} \label{eqn:rank-1 update}
 T(I,v,w) := \sum_{j,l \in [d]} v_j w_l T(:,j,l) \ \in \R^d,
\end{equation}
which is a multilinear combination of the tensor mode-$1$ fibers.
Similarly $T(u,v,w) \in \R$ is a multilinear combination of the tensor entries,  and $T(I, I, w) \in \R^{d \times d}$ is a linear combination of the tensor slices. 

A $3$rd order tensor $T \in \Rbb^{d \times d \times d}$ is said to be rank-$1$ if it can be written in the form
\begin{equation} \label{eqn:rank-1 tensor}
T= \lambda \cdot a \otimes b\otimes c \Leftrightarrow T(i,j,l) = \lambda \cdot a(i) \cdot b(j) \cdot c(l),
\end{equation}
where notation $\otimes$  represents the {\em outer product} and $a, b , c \in \Rbb^d$ are unit vectors.
A tensor $T  \in \Rbb^{d \times d \times d}$ is said to have a CP {\em rank} at most $k$ if it can be written as the sum of $k$ rank-$1$ tensors as
\begin{equation}\label{eqn:tensordecomp}
T = \sum_{i\in [k]} \lambda_i a_i \otimes b_i \otimes c_i, \quad \lambda_i \in \Rbb, \ a_i,b_i,c_i \in \Rbb^d.
\end{equation}

In the rest of the paper, Section~\ref{sec:learning} describes how to apply our tensor results to learning multiview mixture models. Section~\ref{sec:proof-outline} illustrates the proof ideas, with more details in the Appendix. Finally we conclude in Section~\ref{sec:conclusion}.

\section{Learning multiview mixture model through tensor methods} \label{sec:learning}


We proposed our main technical result in Section~\ref{sec:summary} providing convergence guarantees for the tensor power iterations given mild initialization conditions in the overcomplete regime; see Theorem~\ref{thm:dynamics}. Along this result we provide the application to learning multiview mixtures model in Theorem~\ref{thm:guarantees-informal}. In this section, we briefly introduce the tensor decomposition framework as the learning algorithm and then state the learning guarantees with more details and remarks.

\subsection{Multiview mixture model} \label{sec:multiview}

Consider an exchangeable multiview mixture model with $k$ components and $p \geq 3$ views; see Figure \ref{fig:Multiview}. Suppose that hidden variable $h$ is a discrete categorical random variable taking one of the $k$ states. It is convenient to represent it by basis vectors such that $$h = e_j \in \Rbb^k \textnormal{\quad if and only if \quad it takes the $j$-th state.}$$ Note that $e_j  \in \Rbb^k$ denotes the $j$-the basis vector in the $k$-dimensional space. The prior probability for each hidden state is  also $\Pr [h = e_j] = \lambda_j, j \in [k]$. For simplicity, in this paper we assume all the $\lambda_i$'s are the same. However, similar argument works even when the ratio of maximum and minimum prior probabilities $\lambda_{\max} / \lambda_{\min}$ is bounded by some constant.

\begin{figure}
\begin{center}
\begin{tikzpicture}
  [
    scale=1.0,
    observed/.style={circle,minimum size=0.55cm,inner
sep=0mm,draw=black,fill=black!20},
    hidden/.style={circle,minimum size=0.55cm,inner sep=0mm,draw=black},
  ]
  \node [hidden,name=h] at ($(0,0)$) {$h$};
  \node [observed,name=x1] at ($(-1.5,-1)$) {$z_1$};
  \node [observed,name=x2] at ($(-0.5,-1)$) {$z_2$};
  \node [observed,name=xl] at ($(1.5,-1)$) {$z_p$};
  \node at ($(0.5,-1)$) {$\dotsb$};
  \draw [->] (h) to (x1);
  \draw [->] (h) to (x2);
  \draw [->] (h) to (xl);
\end{tikzpicture}
\end{center}
\caption{\small Multiview mixture model.}
\label{fig:Multiview}
\end{figure}

The variables (views) $z_l \in \Rbb^d$ are related to the hidden state through {\em factor matrix} $A  \in \R^{d \times k}$ such that
$$
z_l = A h + \eta_l, \quad l \in [p],
$$
where zero-mean noise vectors $\eta_l \in \R^d$ are independent of each other and the hidden state $h$. Given this, the variables (views) $z_l \in \Rbb^d$ are conditionally independent given the latent variable $h$, and the conditional means are
$\Ebb[z_l|h=e_j] = a_j$, where $a_j \in \R^d$ denotes the $j$-th column of factor matrix $A=[a_1 \dotsb a_k] \in \R^{d \times k}$.
In addition, the above properties imply that the order of observations $z_l$ do not matter and the model is {\em exchangeable}.
The goal of the learning problem is to recover the parameters of the model (factor matrix) $A$ given observations.

For this model, the third order\,\footnote{It is enough to form the third order moment for our learning purpose.} observed moment has the form~\citep{JMLR:v15:anandkumar14b}
\begin{equation}\label{eqn:tensordecompMixture}
\Ebb [z_1 \otimes z_2 \otimes z_3] = \sum_{j \in [k]} \lambda_j a_j \otimes a_j \otimes a_j.
\end{equation}
Hence, given third order observed moment, the unsupervised learning problem (recovering factor matrix $A$) reduces to computing a tensor decomposition as in \eqref{eqn:tensordecompMixture}.

\subsection{Tensor decomposition algorithm} \label{sec:algorithm}

The algorithm for unsupervised learning of multiview mixture model is based on tensor decomposition techniques provided in Algorithm~\ref{algo:Power method form}.
The main step in~\eqref{eqn:power update} performs {\em tensor power iteration}\,\footnote{This is the generalization of matrix power iteration to $3$rd order tensors.}; see~\eqref{eqn:rank-1 update} for the multilinear form definition. 
 After running the algorithm for all different initialization vectors, the clustering process from \citet{AltTensorDecomp2014} ensures that the best converged vectors are returned as the estimation of true components $a_j$.

\begin{algorithm}[t]
\caption{Learning multiview mixture model via tensor power iterations}
\label{algo:Power method form}
\begin{algorithmic}[1]
\REQUIRE 1) Third order moment tensor $T \in \Rbb^{d \times d \times d}$ in~\eqref{eqn:tensordecompMixture}, 2) $n$ samples of $z_1$ in multiview mixture model as $z_1^{(\tau)}, \tau \in [n]$, and 3) number of iterations $N$.
\FOR{$\tau=1$ \TO $n$}
\STATE \textbf{Initialize} unit vectors $x_\tau^{(1)} \leftarrow z_1^{(\tau)} / \bigl\| z_1^{(\tau)} \bigr\|$.
\FOR{$t=1$ \TO $N$}
\STATE Tensor power updates (see \eqref{eqn:rank-1 update} for the definition of the multilinear form):
\begin{equation} \label{eqn:power update}
x_\tau^{(t+1)} = \frac{T \left( I, x_\tau^{(t)}, x_\tau^{(t)} \right)}{\left\| T \left( I, x_\tau^{(t)}, x_\tau^{(t)} \right) \right\|}, \quad
\end{equation}
\ENDFOR
\ENDFOR
\RETURN the output of Procedure~\ref{alg:cluster} with input $\left\{ x_\tau^{(N+1)} : \tau \in [n] \right\}$ as estimates $x_j$.
\end{algorithmic}
\end{algorithm}

\floatname{algorithm}{Procedure}
\begin{algorithm}[t]
\caption{Clustering process~\citep{AltTensorDecomp2014}}
\label{alg:cluster}
\begin{algorithmic}[1]
\REQUIRE Tensor $T \in \Rbb^{d \times d \times d}$, set 
$S := \left\{ x_\tau^{(N+1)}:\tau\in [n]\right\}$, parameter $\nu$.
\WHILE{$S$ is not empty} 
\STATE Choose $x \in S$ which maximizes $|T(x,x,x)|$.
\STATE Do $N$ more iterations of \eqref{eqn:power update} starting from $x$.
\STATE {\bf Output} the result of iterations denoted by $\hx$.
\STATE Remove all the $x \in S$ with $|\langle x, \hx\rangle| > \nu/2$.
\ENDWHILE
\end{algorithmic}
\end{algorithm}

\subsection{Learning guarantees}


We assume a Gaussian prior on the mean vectors, i.e., the vectors $a_j \sim \Nc(0, I_d/d)$, $j \in [k]$ are i.i.d.\ drawn from  a standard multivariate Gaussian distribution with unit expected square norm. Note that in the high dimension (growing $d$), this assumption is the same as uniformly drawing from unit sphere since the norm of vector concentrates in the high dimension and there is no need for normalization.
Even though we impose a prior distribution, we do not use  a MAP estimator, since the corresponding optimization is NP-hard. Instead, we learn the model parameters through decomposition of  the third order moments through   tensor power iterations. The assumption of a Gaussian prior is standard in machine learning applications. We impose it here  for tractable analysis of power iteration dynamics. Such Gaussian assumptions have been used before for analysis of other iterative methods such as  approximate message passing algorithms, and there are evidences that similar results  hold for more general distributions; see \citep{AMP2010} and references there.

As explained in the previous sections, we use tensor power method to learn the components $a_j$'s, and the method is initialized with observed samples $z_i$.
Intuitively, this initialization is useful since $z_i = Ah + \eta_i$ is a perturbed   version of desired parameter $a_j$ (when $h=e_j$). Thus, we present the result in terms of the signal-to-noise (SNR) ratio which is the expected norm of signal $a_j$ (which is one here) divided by the expected norm of noise $\eta_i$, i.e., the SNR in the $i$-th sample $z_i = a_j + \eta_i$ (assumed $h=e_j$) is defined as $\snr := \E[\|a_j\|] / \E[\|\eta_i\|]$. This specifies how much noise the initialization vector $z_i$ can tolerate in order to  ensure the convergence of tensor power iteration to a desired local optimum. 
We now propose the conditions required for recovery guarantees, and state a brief explanation of them.

\paragraph{Conditions for Theorems~\ref{thm:guarantees-informal}~and~\ref{thm:guarantees}:}
\bi[itemsep=-1mm]
\item Rank condition: $k \leq o(d^{1.5})$. 
\item The columns of $A$ are uniformly  i.i.d.\ drawn from unit $d$-dimensional sphere.
\item The noise vectors $\eta_l, l \in [3]$, are independent of matrix $A$ and each other. In addition, the signal-to-noise ratio (SNR) is w.h.p.\ bounded as
$$\snr \geq \Omega \left( \frac{\sqrt{\max\{k,d\}}}{d^{1-\beta}} \right),$$
for some $\beta \geq (\log d)^{-c}$ for universal constant $c>0$.
\end{itemize}

The rank condition bounds the level of overcompleteness for which the recovery guarantees are satisfied. The random assumption on the columns of $A$ are crucial for analyzing the dynamics of tensor power iteration. We use it to argue there exists enough randomness left in the components after conditioning on the previous iterations; see Section~\ref{sec:proofoutline} for the details. The bound on the SNR is required to make sure the given sample used for initialization is close enough to the corresponding mean vector. This ensures that the initial vector is inside the basin-of-attraction of the corresponding component, and hence, the convergence to the mean vector can be guaranteed. Under these assumptions we have.

\begin{theorem}[Learning multiview mixture model given exact tensor: closeness to single columuns] \label{thm:guarantees-informal}
Consider a multiview mixture model (or a spherical Gaussian mixture) in the above setting  with $k$ components in $d$ dimensions. If the above conditions hold, then the tensor power iteration   converges to a vector  close to one of   the true mean vectors $a_j$'s (having constant correlation). 
\end{theorem}

In particular, for mildly overcomplete models, where $k=\alpha d$ for some constant $\alpha>1$, the signal-to-noise ratio  (SNR) is as low as  $\Omega(d^{-1/2+\epsilon})$, for any $\epsilon>0$. Thus, we can learn mixture models with a high level of noise. In general, we establish how the required noise level  scales with the number of hidden components $k$, as long as   $k =o( d^{1.5})$.

The above theorem states convergence to desired local optima which are close to true components $a_j$'s.
In Theorem~\ref{thm:guarantees}, we show that we can sharpen the above result, by jointly iterating over the recovered vectors, and consistently recover the components $a_j$'s. This result also uses the analysis from~\citet{AltTensorDecomp:COLT2015}.

%

\begin{theorem}[Learning multiview mixture model given exact tensor: recovering the whole factor matrix] \label{thm:guarantees}
Assume the above conditions hold. The initialization of power iteration is performed by samples of $z_1$ in multiview mixture model. Suppose the tensor power iterations is at least initialized once for each $a_j, j \in [k]$ such that $z_1 = a_j + \eta_1$.\footnote{Note that this happens for component $j$ with high probability when the number of initializations is proportional to inverse prior probability corresponding to that mixture.} Then by using the exact 3rd order moment tensor in~\eqref{eqn:tensordecompMixture} as input, the tensor decomposition algorithm outputs an estimate $\hA$ satisfying w.h.p.\,(over the randomness of the components $a_j$'s)
$$
 \left\| \hA - A \right\|_F \leq \epsilon,
$$
where the number of iterations of the algorithm is $N = \Theta \left( \log \left( \frac{1}{\epsilon} \right) + \log \log d \right)$.
\end{theorem}


The above theorems assume the exact third order tensor is given to the algorithm. We provide the results given empirical tensor in Section~\ref{sec:sample-complexity}.

\paragraph{Learning spherical Gaussian mixtures:}
Consider a mixture of $k$ different Gaussian vectors with spherical covariance. Let $a_j \in \R^d, j \in [k]$ denote the mean vectors and the covariance matrices are $\sigma^2 I$. Assuming the parameter $\sigma$ is known, the modified third order observed moment 
\[ M_3 := \Ebb[z \otimes z \otimes z] - \sigma^2 \sum_{i \in [d]} \left( \Ebb[z] \otimes e_i \otimes e_i + e_i \otimes \Ebb[z] \otimes e_i + e_i \otimes e_i \otimes \Ebb[z] \right) \]
has the tensor decomposition form \citep{SphericalGaussian2012}
\[M_3 = \sum_{j \in [k]} \lambda_j a_j \otimes a_j \otimes a_j,\]
where $\lambda_j$ is the probability of drawing $j$-th Gaussian mixture. The above guarantees can be applied to learning mean vectors $a_j$ in this model with the additional property that the noise is spherical Gaussian.

\paragraph{Learning multiview mixture model with distinct factor matrices:}
Consider the multiview mixture model with different factor matrices where the first three views are related to the hidden state as
\[z_1 = Ah+\eta_1, \quad z_2 = Bh+\eta_2, \quad z_3 = Ch+\eta_3. \]
Then, the guarantees in the above theorem can be extended to recovering the columns of all three factor matrices $A$, $B$, and $C$ with appropriate modifications in the power iteration algorithm as follows. First the update formula~\eqref{eqn:power update} is changed as 
\begin{equation*} 
x_{1,\tau}^{(t+1)} = \frac{T \left( I, x_{2,\tau}^{(t)}, x_{3,\tau}^{(t)} \right)}{\left\| T \left( I, x_{2,\tau}^{(t)}, x_{3,\tau}^{(t)} \right) \right\|}, \quad
x_{2,\tau}^{(t+1)} = \frac{T \left( x_{1,\tau}^{(t)}, I, x_{3,\tau}^{(t)} \right)}{\left\| T \left( x_{1,\tau}^{(t)}, I, x_{3,\tau}^{(t)} \right) \right\|}, \quad
x_{3,\tau}^{(t+1)} = \frac{T \left( x_{1,\tau}^{(t)}, x_{2,\tau}^{(t)},I \right)}{\left\| T \left( x_{1,\tau}^{(t)}, x_{2,\tau}^{(t)},I \right) \right\|},
\end{equation*}
which is the alternating asymmetric version of symmetric power iteration in~\eqref{eqn:power update}. Here, we alternate among different modes of the tensor. In addition, the initialization for each mode of the tensor is appropriately performed with the samples corresponding to that mode. Note that the analysis still works in the asymmetric version since there exists even more independence relationships through the iterations of the power update because of introducing new random matrices $B$ and $C$.

\subsubsection{Sample complexity analysis} \label{sec:sample-complexity}
In the previous section, we assumed the exact third order tensor in~\eqref{eqn:tensordecompMixture} is given to the tensor decomposition Algorithm~\ref{algo:Power method form}. We now estimate the tensor given $n$ samples $z_1^{(i)},z_2^{(i)},z_3^{(i)}, i \in [n]$, as
\begin{equation} \label{eqn:empirical-tensor}
\hT = \frac{1}{n} \sum_{i \in [n]} z_1^{(i)} \otimes z_2^{(i)} \otimes z_3^{(i)}.
\end{equation}

For the multiview mixture model introduced in Section~\ref{sec:multiview}, let the noise vector $\eta_l$ be spherical, and $\zeta^2$ denote the variance of each entry of noise vector. We now provide the following recovery guarantees.

\paragraph{Additional conditions for Theorem~\ref{thm:guarantees-empirical}:}
\bi[itemsep=-1mm]
\item Let 
$E_1 : = [\eta^{(1)}_1,\eta^{(2)}_1,\dotsc,\eta^{(n)}_1] \in \Rbb^{d \times n},$
where $\eta^{(i)}_1 \in \R^d$ is the $i$-th sample of noise vector $\eta_1$. These noise matrices satisfy the following {\em RIP property} which is adapted from \citet{candes2006near}. Matrix $E_1 \in \Rbb^{d \times n}$ satisfies a weak RIP condition such that for any subset of $O\left(\frac{d}{\log^2 d}\right)$ number of columns, the spectral norm of $E_1$ restricted to those columns is bounded by $2$. The same condition is satisfied for similarly defined noise matrices $E_2$ and $E_3$.
\item The number of samples $n$ satisfies lower bound such that
\begin{equation} \label{eqn:sample-complexity}
\zeta \left(\frac{\sqrt{d}}{n} + \sqrt{\lambda_{\max}\frac{d}{n}}\right) + \zeta^2\left(\frac{d}{n}+\sqrt{\lambda_{\max}\frac{d^{1.5}}{n}}\right)+\zeta^3 \left(\frac{d^{1.5}}{n}+\sqrt{\frac{d}{n}}\right)
\leq \min \left\{ \epsilon \frac{\sqrt{k}}{d}, \tl{O}(\lambda_{\min}) \right\}, 
\end{equation}
where $\epsilon < o \left( \sqrt{k}/d \right)$.
\end{itemize}

\begin{theorem}[Learning multiview mixture model given empirical tensor] \label{thm:guarantees-empirical}
Consider the empirical tensor in \eqref{eqn:empirical-tensor} as the input to tensor decomposition Algorithm~\ref{algo:Power method form}. Suppose the above additional conditions are also satisfied. Then, the same guarantees as in Theorem~\ref{thm:guarantees-informal} hold. In addition, the same guarantees as in Theorem~\ref{thm:guarantees} also hold with the recovery bound changed as
$$
 \left\| \hA - A \right\|_F \leq \tl{O} \left( \frac{\sqrt{k} \cdot \|E\|}{\lambda_{\min}} \right),
$$
where $E$ denotes the perturbation tensor originated from empirical estimation in~\eqref{eqn:empirical-tensor}, and its spectral norm $\|E\|$ is bounded by the LHS of~\eqref{eqn:sample-complexity}.
\end{theorem}

\bprf
The above sample complexity result is proved by using the tensor concentration bound in Theorem 1 of \citet{OvercompleteLVMs2014} applied to our noisy analysis of tensor power dynamics in Theorem~\ref{thm:dynamics}; see Equation~\eqref{eqn:perturbation-bound}. The additional bound on sample complexity and final recovery error on $ \left\| \hA - A \right\|_F$ is also from Theorem 1 of \citet{AltTensorDecomp:COLT2015}.
\eprf

\section{Proof Outline} \label{sec:proof-outline}

Our main technical result is the analysis of third order tensor power iteration provided in Theorem~\ref{thm:dynamics} which also allows to tolerate some amount of noise in the input tensor. 
We analyze the noiseless and noisy settings in different ways. We basically first prove the result for the noiseless setting where the input tensor has an exact rank-$k$ decomposition in~\eqref{eqn:tensordecomp-intro}. When the noise is also considered, we show that the contribution of noise in the analysis is dominated by the main signal, and thus, the same result still holds. For the rest of this section we focus on the noiseless setting, while we discuss the proof ideas for the noisy case in Section~\ref{sec:proofoutline-noise}.

We first discuss the proof of Theorem~\ref{thm:guarantees} which involves two phases. 
In the first phase, we show that under certain small amount of correlation (see \eqref{eqn:init assumption}) between the initial vector and the true component, the power iteration in~\eqref{eqn:poweriteration} converges to some vector which has constant correlation with the true component. This result is the core technical analysis of this paper which is provided in Lemma~\ref{lem:init}.
In the second phase, we incorporate the result of~\citet{AltTensorDecomp2014} which guarantees the approximate convergence of power iteration given initial vector having constant correlation with the true component. This is stated in Lemma~\ref{lem:prevwork}.

To simplify the notation, we consider the tensor\,\footnote{In the analysis, we assume that all the weights are equal to one which can be generalized to the case when the ratio of maximum and minimum weights (in absolute value) are constant.}
%
\begin{equation}
\label{eqn:tensorform}
T= \sum_{j\in [k]} a_j \otimes a_j \otimes a_j, \quad a_j \sim \mathcal{N}(0,\frac{1}{d} I_d).
\end{equation}
Notice that this is exactly proportional to the 3rd order moment tensor of the multiview mixture model in~\eqref{eqn:tensordecompMixture}.


The following lemma is restatement of Theorem~\ref{thm:dynamics} in the noiseless setting.

\begin{lemma}[Dynamics of tensor power iteration, phase 1] \label{lem:init}
Consider the rank-$k$ tensor $T$ of the form in~\eqref{eqn:tensorform}. 
Let tensor rank $k=o(d^{1.5})$, and the unit-norm initial vector $x^{(1)}$ satisfies the correlation bound
\begin{equation}
|\inner{x^{(1)},a_j}| \geq  d^\beta \frac{\sqrt{k}}{d}, \label{eqn:init assumption}
\end{equation}
w.r.t.\ some true component $a_j, j \in [k]$, for some $\beta>(\log d)^{-c}$ for some universal constant $c>0$. After $N = \Theta \left( \log \log d \right)$ iterations, the tensor power iteration in~\eqref{eqn:poweriteration} outputs a vector having w.h.p.\ a constant correlation with the true component $a_j$ as
$$|\inner{x^{(N+1)},a_j}| \ge 1-\gamma,$$ for any fixed constant $\gamma>0$.
\end{lemma}

The proof outline of above lemma is provided in Section~\ref{sec:proofoutline}.

\begin{lemma}[Dynamics of tensor power iteration, phase 2 \citep{AltTensorDecomp2014}] \label{lem:prevwork}
Consider the rank-$k$ tensor $T$ of the form in~\eqref{eqn:tensorform} with rank condition $k \leq o(d^{1.5})$. Let the initial vectors $x^{(1)}_j$ satisfy the constant correlation bound
\begin{equation*}
|\inner{x^{(1)}_j,a_j}| \geq  1-\gamma_j,
\end{equation*}
w.r.t.\ true components $a_j, j \in [k]$, for some constants $\gamma_j>0$. 
Let the output of the tensor power update\,\footnote{This result also needs an additional step of coordinate descent iterations since the true components are not the fixed points of power iteration; see~\citet{ AltTensorDecomp2014} for the details.} in~\eqref{eqn:poweriteration} applied to all these different initialization vectors after $N = \Theta \left( \log \frac{1}{\epsilon} \right)$ iterations be stacked in matrix $\hA$. Then, we have w.h.p.\,\footnote{\citet{AltTensorDecomp2014} recover the vector up to sign since they work in the asymmetric case. In symmetric case it is easy to resolve sign ambiguity issue.}
$$\left\| \hA-A \right\|_F \leq \epsilon.$$
\end{lemma}

Given the above two lemmas, the learning result in Theorem~\ref{thm:guarantees} is directly proved.

\bprfof{Theorem~\ref{thm:guarantees}}
The result is proved by combining Lemma~\ref{lem:init} and Lemma~\ref{lem:prevwork}. Note that the initialization condition in~\eqref{eqn:init} is w.h.p.\ satisfied given the SNR bound assumed.
%
\eprfof

\subsection{Proof outline of Lemma~\ref{lem:init} (noiseless case of Theorem~\ref{thm:dynamics})} \label{sec:proofoutline}

\paragraph{First step:} We first intuitively show the first step of the algorithm makes progress.
 Suppose the tensor is $T=\sum_{j \in [k]} a_j\otimes a_j\otimes a_j$, and the initial vector $x$ has correlation $|\inner{x,a_1}| \ge d^\beta \frac{\sqrt{k}}{d}$ with the first component. The result of the first iteration is the normalized version of the following vector:
$$
\tilx = \sum_{j \in [k]} \inner{a_j,x}^2 a_j.
$$
Intuitively, this vector should have roughly $\inner{a_1,\tilx} = d^{2\beta}\frac{k}{d^2}$ correlation with $a_1$ (as the other terms are random they don't contribute much). On the other hand, the norm of this vector is roughly $O(\sqrt{k}/d)$: this is because $\inner{a_j,x}^2$ for $j\ne 1$ is roughly\,\footnote{The correlation between two unit Gaussian vectors in  $d$ dimensions is roughly $1/\sqrt{d}$.} $1/d$, and the sum of $k$ random vectors with length $1/d$ will have length roughly $O(\sqrt{k}/d)$. These arguments can be made precise showing the normalized version $\tilx/\| \tilx\|$ has correlation $d^{2\beta}\frac{\sqrt{k}}{d}$ with $a_1$ ensuring progress in the first step.

\paragraph{Going forward:} As we explained, the basic idea behind proving Lemma~\ref{lem:init} is to characterize the conditional distribution of random Gaussian tensor components $a_j$'s given previous iterations. In particular, we show that the residual independent randomness left in these conditional distributions is large enough and we can exploit it to obtain tighter concentration bounds throughout the analysis of the iterations. The Gaussian assumption on the components, and small enough number of iterations are crucial in this argument.


\paragraph{Notations:} For two vectors $u,v \in \R^k$, the Hadamard product denoted by $*$ is defined as the entry-wise multiplication of vectors, i.e., $(u*v)_j := u_jv_j$ for $j \in [k]$.
For a matrix $A$, let  $P_{\perp_A}$ denote the projection operator to the subspace orthogonal to column span of $A$. For a subspace $R$, let $R^\perp$ denote the space orthogonal to it. Therefore, for a subspace $R$, the projection operator on the subspace orthogonal to $R$ is equivalently denoted by $P_{R^\perp}$ or $P_{\perp_R}$.
For a random matrix $D$, let $D|\{u = Dv\}$ denote the conditional distribution of $D$ given linear constraints $u = Dv$.

Lemma~\ref{lem:init} involves analyzing the dynamics of power iteration in~\eqref{eqn:poweriteration} for 3rd order rank-$k$ tensors. For the rank-$k$ tensor in~\eqref{eqn:tensorform}, the power iterative form $x \leftarrow \frac{T(I,x,x)}{\|T(I,x,x)\|}$ can be written as
\begin{equation}
x^{(t+1)} = \frac{A \left( A^\top x^{(t)} \right)^{*2}}{ \left\|A \left( A^\top x^{(t)} \right)^{*2} \right\|}, \label{eqn:update formula}
\end{equation}
where the multilinear form in~\eqref{eqn:rank-1 update} is used. Here, $A = [a_1 \dotsb a_k] \in \R^{d \times k}$ denotes the factor matrix, and for vector $y \in \R^k$, $y^{*2}:= y*y \in \R^k$ represents the element-wise square of entries of $y$. 


%

We consider the case where $a_i \sim \Nc(0, \frac{1}{d}I)$ are i.i.d.\ drawn and we analyze the evolution of the dynamics of the power update. As explained earlier, for a given initialization $x^{(1)}$, the update in the first step can be analyzed easily since $A$ is independent of $x^{(1)}$. However, in subsequent steps, the updates $x^{(t)}$ are dependent on $A$, and it is no longer clear how to provide a tight bound on the evolution of $x^{(t)}$. In this work, we provide a careful analysis by controlling the amount of ``correlation build-up'' by exploiting the structure of Gaussian matrices under linear constraints. This enables us to provide better guarantees for matrix $A$ with Gaussian entries compared to general matrices $A$.

\paragraph{Intermediate update steps and variables:}
Before we proceed, we need to break down power update in~\eqref{eqn:poweriteration} and introduce some intermediate update steps and variables as follows. Recall that $x^{(1)} \in \R^d$ denotes the initialization vector.
Without loss of generality, let us analyze the convergence of power update to first component of rank-$k$ tensor $T$ denoted by $a_1$. Hence, let the first entry of $x^{(1)}$ denoted by $x_1^{(1)}$ be the maximum entry (in absolute value) of $x^{(1)}$, i.e., $x_1^{(1)}=\|x^{(1)}\|_\infty$.
Let $B:=[a_2 \ a_3 \ \dotsb \ a_k] \in \R^{d \times (k-1)}$, and therefore $A = [a_1 | B]$. We break the power update formula in~\eqref{eqn:poweriteration} into a few steps by introducing intermediate variables $y^{(t)} \in \R^k$ and $\tilx^{(t+1)} \in \R^d$ as
\[
y^{(t)} := A^\top x^{(t)}, \quad \tilx^{(t+1)} := A  (y^{(t)})^{*2}.
\]
Note that $\tilx^{(t+1)}$ is the unnormalized version of $x^{(t+1)} :=\tilx^{(t+1)} / \|\tilx^{(t+1)}\|$, i.e., $\tilx^{(t+1)} := T(I,x^{(t)},x^{(t)})$. 
Thus, we need to jointly analyze the dynamics of all variables $x^{(t)}$,  $y^{(t)}$ and $(y^{(t)})^{*2}$.
Define \[  X^{[t]}:= \left[ x^{(1)}| \ldots| x^{(t)} \right], \quad Y^{[t]}:= \left[ y^{(1)}| \ldots| y^{(t)} \right].\]
Matrix $B$ is randomly drawn with i.i.d.\ Gaussian entries $B_{ij} \sim \Nc(0, \frac{1}{d})$. As the iterations proceed, we consider the following conditional distributions
\begin{equation}
B^{(t,1)} := B|\{X^{[t]}, Y^{[t]} \}, \quad B^{(t,2)}  := B| \{X^{[t+1]}, Y^{[t]} \}.\label{eqn:Bdef}
\end{equation}
Thus, $B^{(t,1)}$ is the conditional distribution of $B$ at the middle of $t^{\tha}$ iteration (before update step $\tilx^{(t+1)} = A (y^{(t)})^{*2}$) and $B^{(t,2)}$ is the conditional distribution at the end of $t^{\tha}$ iteration (after update step $\tilx^{(t+1)} = A (y^{(t)})^{*2}$). 
By analyzing the above conditional distributions, we can characterize the left independent randomness in $B$.


\subsubsection{Conditional Distributions}

In order to characterize the conditional distribution of $B$ under evolution of $x^{(t)}$ and $y^{(t)}$ in \eqref{eqn:Bdef}, we exploit the following basic fact (see~\citep{AMP2010} for proof).

\begin{lemma}[Conditional distribution of Gaussian matrices under a linear constraint]\label{lemma:cond} Consider random matrix $D$ with i.i.d.\ Gaussian entries $D_{ij} \sim \Nc(0, \sigma^2)$. Conditioned on $u = Dv$ with known vectors $u$ and $v$, the matrix $D$ is distributed as
 \[ D|\{u =Dv\}\eqd \frac{1}{\norm{v}^2}uv^\top + \tilde{D} P_{\perp_v}, \]
 where random matrix $\tilde{D}$ is an independent copy of $D$ with i.i.d.\ Gaussian entries $\tl{D}_{ij} \sim \Nc(0, \sigma^2)$, and $P_{\perp_v}$ is the projection operator on to the subspace orthogonal to $v$.
\end{lemma}

We refer to $\tilde{D} P_{\perp_v}$ as the   {\em residual} random matrix  since it represents the remaining {\em randomness} left after conditioning. It   is  a random matrix whose rows are independent random vectors that are orthogonal to $v$, and the variance in each direction orthogonal to $v$ is equal to $\sigma^2$.

The above Lemma can be exploited to characterize the conditional distribution of $B$ introduced in \eqref{eqn:Bdef}. However, a naive direct application using the constraint $Y^{[t]}= A^\top X^{[t]}$ is not transparent for analysis. The reason is  the evolution of $x^{(t)}$ and $y^{(t)}$ are themselves governed by the conditional distribution of $B$ given previous iterations. Therefore, we need the following recursive version of Lemma~\ref{lemma:cond}.

\begin{corollary}[Iterative conditioning]\label{corr:itercond}
Consider random matrix $D$ with i.i.d.\ Gaussian entries $D_{ij} \sim \Nc(0, \sigma^2)$, and let $F\eqd P_{\perp_C} D P_{\perp_R}$ be the  random Gaussian matrix whose columns are orthogonal to space $C$ and rows are orthogonal to space  $R$. Conditioned on the linear constraint $u = Dv$, 
where\,\footnote{We need that $u \in C^\perp$, otherwise the event $u=Dv$ is impossible.} $u \in C^\perp$, 
the matrix $F$ is distributed as \[F|\{u=Dv\}\eqd \frac{1}{\norm{(P_{\perp_R} v)}^2}u ( P_{\perp_R} v)^\top+P_{\perp_C}\tilde{D} P_{\perp_{\{R, v\}}},\] where random matrix $\tilde{D}$ is an independent copy of $D$ with i.i.d.\ Gaussian entries $\tl{D}_{ij} \sim \Nc(0, \sigma^2)$.
\end{corollary}

Thus, the {\em residual}  random  matrix $P_{\perp_C}\tilde{D} P_{\perp_{\{R, v\}}}$ is a random Gaussian matrix whose columns are orthogonal to $C$ and rows are orthogonal to $\spn\{R,v\}$. The variance in any remaining dimension is equal to $\sigma^2$.

\subsubsection{Form of Iterative Updates}

Now we exploit the conditional distribution arguments proposed in the previous section to characterize the conditional distribution of $B$ given the update variables $x$ and $y$ up to the current iteration; recall~\eqref{eqn:Bdef} where $B^{(t,1)}$ is the conditional distribution of $B$ at the middle of $t^{\tha}$ iteration and $B^{(t,2)}$ at the end of $t^{\tha}$ iteration. Before that, we need to introduce some more intermediate variables.

\paragraph{Intermediate variables:}
We separate the first entry of $y$ and $(y)^{*2}$ from the rest, i.e., we have
\[ y^{(t)}_1= a_1^\top x^{(t)}, \quad y^{(t)}_{-1}=B^\top x^{(t)} \sim (B^{(t-1,2)})^\top x^{(t)}, \]
 where $y^{(t)}_{-1} \in \Rbb^{k-1}$ denotes $y^{(t)} \in \Rbb^k$ with the first entry removed. The update formula for $\tilx^{(t+1)}$ can be also decomposed as
\[ \tilx^{(t+1)} = (y_1^{(t)})^2 a_1 + B w^{(t)} \sim (y_1^{(t)})^2 a_1 + B^{(t,1)} w^{(t)},\]
where
$$w^{(t)}:= (y_{-1}^{(t)})^{*2} \in \Rbb^{k-1},$$
is the new intermediate variable in the power iterations.
 Let $B_{\res}^{(t,1)}$ and $B_{\res}^{(t,2)}$ denote the {\em residual} random matrices corresponding to $B^{(t,1)}$ and $B^{(t,2)}$ respectively, and
\begin{align*}
u^{(t+1)} := B_{\res}^{(t,1)} w^{(t)} ,\quad v^{(t)} := (B^{(t-1,2)}_{\res})^\top x^{(t)},
\end{align*}
where $u^{(t)} \in \R^d$ and $v^{(t)} \in \Rbb^{k-1}$ are respectively the part of $x^{(t)}$ and $y^{(t)}_{-1}$ representing the residual randomness after conditioning on the previous iterations.
We also summarize all variables and notations in Table~\ref{table:notations} in the Appendix which can be used as a reference throughout the paper.

Finally we make the following observations.

\begin{lemma}[Form of iterative updates]\label{lemma:iter}The conditional distribution of $B$ at the middle of $t^{\tha}$ iteration denoted by $B^{(t,1)}$ satisfies
\begin{align}
B^{(t,1)} &\eqd \sum_{i \in [t-1]} \frac{ u^{(i+1)} (P_{\perp_{W^{[i-1]}}} w^{(i)})^\top }{\|P_{\perp_{W^{[i-1]}}} w^{(i)}\|^2}
+ \sum_{i \in [t]} \frac{ P_{\perp_{X^{[i-1]}}} x^{(i)} (v^{(i)})^\top }{\|P_{\perp_{X^{[i-1]}}} x^{(i)}\|^2}
+ B_{\res}^{(t,1)}, \label{eqn:B1} \\
B_{\res}^{(t,1)} & \eqd P_{\perp_{X^{[t]} } } \tilde{B} P_{\perp_{W^{[t-1]}}}, \label{eqn:B_res1}
\end{align}
where random matrix $\tilde{B}$ is an independent copy of $B$ with i.i.d.\ Gaussian entries $\tl{B}_{ij} \sim \Nc(0, \frac{1}{d})$. Similarly, the conditional distribution of $B$ at the end of $t^{\tha}$ iteration denoted by $B^{(t,2)}$ satisfies
\begin{align}
B^{(t,2)} &\eqd  \sum_{i \in [t]} \left( \frac{ u^{(i+1)} (P_{\perp_{W^{[i-1]}}} w^{(i)})^\top }{\|P_{\perp_{W^{[i-1]}}} w^{(i)}\|^2}
+ \frac{ P_{\perp_{X^{[i-1]}}} x^{(i)} (v^{(i)})^\top }{\|P_{\perp_{X^{[i-1]}}} x^{(i)}\|^2} \right)
+ B_{\res}^{(t,2)}, \label{eqn:B2} \\
B_{\res}^{(t,2)} & \eqd P_{\perp_{X^{[t]} } } B' P_{\perp_{W^{[t]}} }, \label{eqn:B_res2}
\end{align}
where random matrix $B'$ is an independent copy of $B$ with i.i.d.\ Gaussian entries $B'_{ij} \sim \Nc(0, \frac{1}{d})$.
\end{lemma}

The lemma can be directly proved by applying the iterative conditioning argument in Corollary~\ref{corr:itercond}. See the detailed proof in the appendix.

\subsubsection{Analysis of Iterative Updates} \label{sec:inductionhyp}

Lemma~\ref{lemma:iter} characterizes the conditional distribution of $B$ given the update variables $x$ and $y$ up to the current iteration; see~\eqref{eqn:Bdef} for the definition of conditional forms of $B$ denoted by $B^{(t,1)}$ and $B^{(t,2)}$. Intuitively, when the number of iterations $t \ll d$, then the residual independent randomness left in $B^{(t,1)}$ and $B^{(t,2)}$ (respectively denoted by $B^{(t,1)}_{\res}$ and $B^{(t,2)}_{\res}$) characterized in Lemma~\ref{lemma:iter} is large enough and we can exploit it to obtain tighter concentration bounds throughout the analysis of the iterations.


Note that the goal is to show that under $t \ll d$, the iterations $x^{(t)}$ converge to the true component with constant error, i.e.,  $|\inner{x^{(t)},a_1}| \ge 1-\gamma$ for some constant $\gamma>0$. If this already holds before iteration $t$ we are done, and if it does not hold, next iteration is analyzed to finally achieve the goal. This analysis is done via {\em induction argument}.
During the iterations, we maintain several invariants to analyze the dynamics of power update. The goal is to ensure progress in each iteration as in \eqref{eqn:progress}.


\begin{figure}[t]
\begin{center}
\begin{tikzpicture}
  [
    scale=1.0,
    operator/.style={circle,minimum size=0.4cm,inner sep=0mm,draw=black},
  ]
    \node at ($(0,0)$) {$\cdots \rightarrow x^{(t)} \longrightarrow
y^{(t)} \longrightarrow
w^{(t)} \longrightarrow
x^{(t+1)} \longrightarrow
y^{(t+1)} \rightarrow \cdots$};
  \draw[dashed] ($(-2.9,0.7)$) to ($(-2.9,-0.9)$);
  \draw[dashed] ($(1.75,0.7)$) to ($(1.75,-0.9)$);
  \draw[<->] ($(-2.9,-0.6)$) to ($(1.75,-0.6)$);
  \node at ($(-0.6,-1.0)$) {update steps at iteration $t$};
 \end{tikzpicture}
\end{center}
\caption{\small Flow of the power update algorithm stating intermediate steps. Iteration $t$ for which the inductive step should be argued is also indicated.}
\label{fig:algorithmflow}
\end{figure}

\paragraph{Induction hypothesis:} The following are assumed at the beginning of the iteration $t$ as induction hypothesis; see Figure~\ref{fig:algorithmflow} for the scope of inductive step.
\begin{enumerate}
\item {\bf Length of Projection on $x$: } \label{hyp:proj_x}
$$\delta_t \le \|P_{\perp_{X^{[t-1]}}} x^{(t)} \| \leq 1,$$
where $\delta_t$ is of order $1/\polylog d$, and the value of $\delta_t$ only depends on $t$ and $\log d$.
\item {\bf Length of Projection on $w$: }\label{hyp:proj_w}
\begin{align*}
\delta'_{t-1} \frac{\sqrt{k}}{d} \le \| P_{\perp_{W^{[t-2]} }} w^{(t-1)}\| & \leq \Delta'_{t-1} \frac{\sqrt{k}}{d}, \\
\| P_{\perp_{W^{[t-2]} }} w^{(t-1)}\|_\infty & \le \Delta'_{t-1} \frac{1}{d},
\end{align*}
where $\delta'_t$ is of order $1/\polylog d$ and $\Delta'_t$ is of order $\polylog d$. Both $\delta'_t$ and $\Delta'_t$ only depend on $t$ and $\log d$.
\item {\bf Progress:}\,\footnote{Note that although the bounds on $y^{(t)}_{-1}$ are argued at iteration $t$, the bound on the first entry of $y^{(t)}$ denoted by $y^{(t)}_1 = \inner{a_1,x^{(t)}}$ is assumed here in the induction hypothesis at the end of iteration $t-1$.}  \label{hyp:prog}
\begin{align}
|\inner{a_1,x^{(t)}}|& \in [\delta^*_t, \Delta^*_t] d^{\beta 2^{t-1}} \frac{\sqrt{k}}{d}, \label{eqn:progress} \\
\inner{a_1,P_{\perp_{X^{[t-1]}}} x^{(t)}} & \le \Delta^*_t d^{\beta 2^{t-1}} \frac{\sqrt{k}}{d}. \nn
\end{align}
\item {\bf Norm of $u$,$v$: } \label{hyp:normuv}
\begin{align*}
\frac{\delta_{t-1}}{2}\sqrt{\frac{k}{d}} & \le \|v^{(t-1)}\| \le 2 \sqrt{\frac{k}{d}}, \\
\frac{\delta'_{t-1}}{2} \frac{\sqrt{k}}{d} & \le \|u^{(t)}\| \le 2\Delta'_{t-1} \frac{\sqrt{k}}{d}.
\end{align*}

\end{enumerate}

The analysis for basis of induction and inductive step are provided in Appendix~\ref{appendix:inductionanalysis}.

\subsection{Effect of noise in Theorem~\ref{thm:dynamics}} \label{sec:proofoutline-noise}

Given rank-$k$ random tensor $T$ in~\eqref{eqn:tensorform}, and a starting point $x^{(1)}$, our  analysis in the  noiseless setting shows that the tensor power iteration in~\eqref{eqn:poweriteration} outputs a vector which will be close to $a_j$ if $x^{(1)}$ has a large enough correlation with $a_j$.

Now suppose we are given noisy tensor $\hat{T} = T+E$ where $E$ has some small norm. In this case where the noise is also present, we get a sequence $\hat{x}^{(t)} = x^{(t)} + \xi^{(t)}$ where $x^{(t)}$ is the  component not incorporating any noise (as in previous section\footnote{Note that there is a subtle difference between notation $x^{(t)}$ in the noiseless and noisy settings. In the noiseless setting, this vector is normalized, while in the noisy setting the whole vector $\hat{x}^{(t)} = x^{(t)} + \xi^{(t)}$ is normalized.}), while $\xi^{(t)}$ represents the contribution of noise tensor $E$ in the power iteration; see~\eqref{eqn:poweriteration-noisy} below. We prove that $\xi^{(t)}$ is a very small noise that does not change our calculations stated in the following lemma.

\begin{lemma} [Bounding norm of error] \label{lem:noise-induction}
Suppose the spectral norm of the error tensor $E$ is bounded as  
$$\|E\| \leq \epsilon \sqrt{k}/d, \quad \textnormal{where} \quad \epsilon < o(\sqrt{k}/d).$$
Then the noise vector $\xi^{(t)}$ at iteration $t$ satisfies the $\ell_2$ norm bound
$$\|\xi^{(t)}\| \leq \tilde{O}(d^{\beta2^{t-1}}\epsilon).$$
Note that when $t$ is the first number such that $d^{\beta2^{t-1}} \geq d/\sqrt{k}$, we have $\|\xi^{(t)}\| = o(1)$.
\end{lemma}

Notice that since when $d^{\beta2^{t-1}} \ge d/\sqrt{k}$, the main induction is already over and we know $x^{(t)}$ is constant close to the true component,  and thus, the noise is always small. 

\paragraph{Proof idea:}
We now provide an overview of ideas for proving the above lemma; see Appendix~\ref{appendix:noise} for the complete proof which is based on an induction argument.
We first write the following recursion expanding the contribution of main signal and noise terms in the tensor power iteration as
\begin{align} 
x^{(t+1)} + \xi^{(t+1)} & = \normalize \left( \hat{T}(x^{(t)} + \xi^{(t)},x^{(t)} + \xi^{(t)},I) \right) \nn \\
&= \normalize \left( T(x^{(t)},x^{(t)},I) + 2T(x^{(t)}, \xi^{(t)}, I) + T(\xi^{(t)},\xi^{(t)},I) + E (\hat{x}^{(t)},\hat{x}^{(t)},I) \right), \label{eqn:poweriteration-noisy}
\end{align}
where for vector $v$, we have $\normalize (v) := v/\|v\|$, i.e., it normalizes the vector. 
The first term is the desired main signal and should have the largest norm, and the rest of the terms are the noise terms.
The third term is of order $\|\xi^{(t)}\|^2$, and hence, it should be fine whenever we choose $\|E\|$ to be small enough. The last term is $O(\|E\|)$ and is the same for all iterations so that is also fine. The problematic term is the second term, whose norm if we bound naively is $2\|\xi^{(t)}\|$. However the normalization factor also contributes a factor of roughly $d/\sqrt{k}$, and thus, this term grows exponentially; it is still fine if we just do a constant number of iterations, but the exponent will depend on the number of iterations.

In order to solve this problem, and make sure that the amount of noise we can tolerate is independent of the number of iterations, we need a better way to bound the noise term $\xi^{(t)}$. 
The main problem here is we bound the norm of $\|T(x^{(t)}, \xi^{(t)}, I)\|$ by $\|T\| \|\xi^{(t)}\| \leq O(\xi^{(t)})$, by doing this we ignored the fact that $x^{(t)}$ is uncorrelated with the components in $T$. In order to get a tighter bound, we introduce another norm $\|\cdot \|_*$. Intuitively, the norm $\|\cdot \|_*$ captures the fact that $x$ does not have a high correlation with the components (except for the first component that $x$ will converge to), and gives a better bound. In particular we have $\|T(x^{(t)}, \xi^{(t)}, I)\| \approx \frac{\sqrt{k}}{d} \|\xi^{(t)}\|_2$.  Therefore, the normalization factor is compensated by the additional term $\frac{\sqrt{k}}{d}$.
More concretely, this norm is defined as follows.


\begin{definition} [Norm $\|\cdot\|_*$] \label{def:starnorm}
Given a matrix $A = [a_1 \ a_2 \ \dotsb \ a_k] \in \R^{d\times k}$, for any vector $u\in \R^d$, the norm $\|u\|_{A^*}$ is defined as $$\|u\|_{A^*} = \max_{i \in [k]} |\inner{a_i,u}|.$$
\end{definition}
 
This norm satisfies a property shown in Lemma~\ref{lem:mixednorm} which enables us to argue that $\xi^{(t)}$ is small enough as stated in Lemma~\ref{lem:noise-induction}.

\section{Conclusion} \label{sec:conclusion}
In this paper, we provide a novel analysis for the dynamics of third order tensor power iteration showing convergence guarantees to vectors having constant correlation with the tensor component. This enables us to prove unsupervised learning of latent variable models in the challenging overcomplete regime where the hidden dimensionality is larger than the observed dimension. The main technical observation is that under random Gaussian tensor components and small number of iterations, the residual randomness in the components (which are involved in the iterative steps) are sufficiently large. This enables us to show progress in the next iteration of the update step. As future work, it is very interesting to generalize this analysis to higher order tensor power iteration, and more generally to other kinds of iterative updates.

\subsubsection*{Acknowledgements}
A. Anandkumar is supported in part by Microsoft Faculty Fellowship, NSF Career award CCF-$1254106$, NSF Award CCF-$1219234$, ARO YIP Award W$911$NF-$13$-$1$-$0084$ and ONR Award N00014-14-1-0665. M. Janzamin is supported by NSF Award CCF-1219234.

\renewcommand{\appendixpagename}{Appendix}

\appendixpage

\appendix

\begin{table}[H]
\begin{center}
\caption{Table of parameters and variables. Superscript $(t)$ denotes the variable at $t$-th iteration.}
\label{table:notations}
{\renewcommand{\arraystretch}{1.5}
\begin{tabular}{|c|c|L{8cm}|l|}
\hline
Variable & Space & Description & Recursion formula \\
\hline \hline
$A$ & $\Rbb^{d \times k}$ & mapping matrix in update formula~\eqref{eqn:update formula} & n.a.  \\
\hline
$x^{(t)}$ & $\Rbb^d$ & update variable in \eqref{eqn:update formula} & $x^{(t+1)} := \frac{A  (y^{(t)})^{*2}}{\|A  (y^{(t)})^{*2}\|}$ \\
$y^{(t)}$ & $\Rbb^k$ & intermediate variable in update formula \eqref{eqn:update formula} & $y^{(t)} := A^\top x^{(t)}$ \\
$\tilx^{(t)}$ & $\Rbb^d$ & unnormalized version of $x^{(t)}$ & $\tilx^{(t+1)} := A  (y^{(t)})^{*2}$ \\
\hline
$\hx^{(t)}$ & $\Rbb^d$ & noisy version of $x^{(t)}$ & $\hat{x}^{(t)} = x^{(t)} + \xi^{(t)}$; see~\eqref{eqn:poweriteration-noisy} \\
$\xi^{(t)}$ & $\Rbb^d$ & Contribution of noise in tensor power update given noisy tensor $\hat{T} = T+E$ & $\hat{x}^{(t)} = x^{(t)} + \xi^{(t)}$; see~\eqref{eqn:poweriteration-noisy} \\
\hline
$B$ & $\Rbb^{d \times (k-1)}$ & matrix $A := [a_1 \ a_2 \ \dotsb \ a_k]$ with first column removed, i.e., $B:=[a_2 \ a_3 \ \dotsb \ a_k]$. Note that the first column $a_1$ is the desired one to recover. & n.a. \\ \hdashline
$B^{(t,1)}$ & $\Rbb^{d \times (k-1)}$ & conditional distribution of $B$ given previous iterations at the middle of $t^{\tha}$ iteration (before update step $\tilx^{(t+1)} = A (y^{(t)})^{*2}$) & $B^{(t,1)} \eqd B|\{X^{[t]}, Y^{[t]} \}$  \\
\hdashline
$B^{(t,2)}$ & $\Rbb^{d \times (k-1)}$ & conditional distribution of $B$ given previous iterations at the end of $t^{\tha}$ iteration (after update step $\tilx^{(t+1)} = A (y^{(t)})^{*2}$) & $B^{(t,2)}  \eqd B| \{X^{[t+1]}, Y^{[t]} \}$ \\
\hdashline
$B_{\res}^{(t,1)}$ & $\Rbb^{d \times (k-1)}$ & residual independent randomness left in $B^{(t,1)}$; see Lemma~\ref{lemma:iter}. & see equation~\eqref{eqn:B_res1} \\
\hdashline
$B_{\res}^{(t,2)}$ & $\Rbb^{d \times (k-1)}$ & residual independent randomness left in $B^{(t,2)}$; see Lemma~\ref{lemma:iter}. & see equation~\eqref{eqn:B_res2} \\
\hline
$w^{(t)}$ & $\Rbb^{k-1}$ & intermediate variable in update formula \eqref{eqn:update formula} & $w^{(t)}:= (y_{-1}^{(t)})^{*2}$ \\
$u^{(t)}$ & $\Rbb^d$ & part of $x^{(t)}$ representing the left independent randomness & $u^{(t+1)} := B_{\res}^{(t,1)} w^{(t)}$ \\
$v^{(t)}$ & $\Rbb^{k-1}$ & part of $y^{(t)}_{-1}$ representing the left independent randomness & $v^{(t)} := (B^{(t-1,2)}_{\res})^\top x^{(t)}$ \\
\hline
\end{tabular}}
\end{center}
\end{table}

\bprfof{Lemma~\ref{lemma:iter}}
Recall that we have updates of  the form\[\tilx^{(t+1)} = A (y^{(t)})^{*2},\quad w^{(t)}:= (y^{(t)}_{-1})^{*2},  \quad y^{(t)}= A^\top x^{(t)}.\]
Let
\[ X^{[t] \setminus 1} := \left[ x^{(2)}| \ldots| x^{(t)} \right], \]
and let the rows of $Y^{[t]}$ are partitioned as the first and the rest of rows as
\[Y^{[t]} = \left[ {Y^{[t]}_1}^\top \Big| {Y^{[t]}_{-1}}^\top \right]^\top.\]

We now make the following simple observations
\begin{align*}
B^{(t,1)} \eqd B|\{& Y^{[t]} = A^\top X^{[t]}, \, \tilX^{[t] \setminus 1}= A (Y^{[t-1]})^{*2}\} \\
\eqd B|\{& Y^{[t]}_{-1} = B^\top X^{[t]}, \, \tilX^{[t]  \setminus 1}= a_1 (Y_1^{[t-1]})^{*2} + B W^{[t-1]}\} \\
\eqd B|\{& v^{(1)} = B^\top x^{(1)}, \dotsc, v^{(t)}= (B_{\res}^{(t-1,2)})^\top x^{(t)}, \\  &u^{(2)} = B_{\res}^{(1,1)} w^{(1)}, \dotsc, u^{(t)}= B_{\res}^{(t-1, 1) }w^{(t-1)} \},
\end{align*} where the second equivalence comes from the fact that $B$ is matrix $A$ with first column removed. Now applying Corollary~\ref{corr:itercond}, we have the result. The distribution of $B^{(t,2)}$ follow similarly.
\eprfof

\section{Analysis of Induction Argument} \label{appendix:inductionanalysis}

In this section, we analyze the basis of induction and inductive step for the induction argument proposed in Section~\ref{sec:inductionhyp} for the proof of Lemma~\ref{lem:init}.

\subsection{Basis of induction}

We first show that the hypothesis holds for initialization vector $x^{(1)}$ as the basis of induction.

\begin{claim}[Basis of induction]
The induction hypothesis is true for $t=1$.
\end{claim}

\bprf
Notice that induction hypothesis for $t=1$ only involves the bounds on $\|x^{(1)}\|$ and $\inner{a_1,x^{(1)}}$ as in Hypotheses~\ref{hyp:proj_x}~and~\ref{hyp:prog}, respectively. These bounds are directly argued by the correlation assumption on the initial vector $x^{(1)}$ stated in~\eqref{eqn:init assumption} where $\delta_1 =  \delta^*_1 = \Delta^*_1 = 1$.
\eprf

\subsection{Inductive step} \label{sec:inductivestep}

Assuming the induction hypothesis holds for all the values till the end of iteration $t-1$ (stated in Section~\ref{sec:inductionhyp}), we analyze the $t$-th iteration of the algorithm, and prove that induction hypothesis also holds for the values at the end of iteration $t$. See Figure~\ref{fig:algorithmflow} where the scope of iteration $t$ and the flow of the algorithm is shown. In the rest of this section, we pursue the flow of the algorithm at iteration $t$ starting from computing $y^{(t)}$ and ending up with computing $x^{(t+1)}$ to prove the desired induction hypothesis at iteration $t$.

\subsubsection*{Hypothesis~\ref{hyp:normuv}}

We start by showing that the induction Hypothesis~\ref{hyp:normuv} holds at iteration $t$ using the induction Hypotheses~\ref{hyp:proj_x}~and~\ref{hyp:proj_w} in the previous iteration.

\begin{claim} We have
\begin{align*}
\frac{\delta_{t}}{2}\sqrt{\frac{k}{d}} & \le \|v^{(t)}\| \le 2 \sqrt{\frac{k}{d}}, \\
\frac{\delta'_{t}}{2} \frac{\sqrt{k}}{d} & \le \|u^{(t+1)}\| \le 2\Delta'_{t} \frac{\sqrt{k}}{d}.
\end{align*}
\end{claim}

\bprf
Recall that $v^{(t)} := (B^{(t-1,2)}_{\res})^\top x^{(t)}$, and by applying the form of $B^{(t-1,2)}_{\res}$ in~\eqref{eqn:B_res2}, we have
\begin{equation} \label{eqn:vt_expansion}
v^{(t)} \eqd P_{\perp_{W^{[t-1]}}} B'^\top P_{\perp_{X^{[t-1]} } } x^{(t)}.
\end{equation}
Since random matrix $B' \in \R^{d \times (k-1)}$ is an independent copy of $B$ with i.i.d.\ Gaussian entries $B'_{ij} \sim \Nc(0, \frac{1}{d})$, we know $v^{(t)}$ is a random Gaussian vector in the subspace orthogonal to $W^{[t-1]}$.
On the other hand, for any vector $z \in \R^d$, we have
$$
\E \left[ \| B'^\top z\|^2 \right] = z^\top \E \left[ B' B'^\top \right] z = \frac{k-1}{d} \|z\|^2,
$$
where $\E \left[ B' B'^\top \right] = \frac{k-1}{d} I$ is exploited. Let $z = P_{\perp_{X^{[t-1]} } } x^{(t)}$. Then, by applying the above equality to the expansion of $v^{(t)}$ in~\eqref{eqn:vt_expansion}, we have 
$$
\E \left[ \| v^{(t)} \|^2 \right] = \frac{k-t}{k-1} \cdot \frac{k-1}{d} \cdot \|P_{\perp_{X^{[t-1]} } } x^{(t)}\|^2 
= \frac{k-t}{d} \cdot \|P_{\perp_{X^{[t-1]} } } x^{(t)}\|^2 
\in \left[ \delta_t^2 \frac{k}{d} \left(1-\frac{t}{k}\right), \frac{k}{d} \right], 
$$
where $\dim(W^{[t-1]}) = t-1$ is also used in the first step, and the last step is concluded from Hypothesis~\ref{hyp:proj_x}. Finally, by concentration property of random Gaussian vectors, when $t\ll d$ we have with high probability $$\|v^{(t)}\| \in \left[ \frac{\delta_t}{2} \sqrt{\frac{k}{d}},  2\sqrt{\frac{k}{d}} \right].$$


Similarly, for $u^{(t+1)} := B^{(t,1)}_{\res} w^{(t)}$, and by applying the form of $B^{(t,1)}_{\res}$ in~\eqref{eqn:B_res1}, we have
\begin{equation} \label{eqn:ut_expansion}
u^{(t+1)} \eqd P_{\perp_{X^{[t]} } } \tilde{B} P_{\perp_{W^{[t-1]}}} w^{(t)}.
\end{equation}
Since random matrix $\tilde{B} \in \R^{d \times (k-1)}$ is an independent copy of $B$ with i.i.d.\ Gaussian entries $\tilde{B}_{ij} \sim \Nc(0, \frac{1}{d})$, we know $u^{(t+1)}$ is a random Gaussian vector in the subspace orthogonal to $X^{[t]}$.
On the other hand, for any vector $z \in \R^{k-1}$, we have
$$
\E \left[ \| \tilde{B} z\|^2 \right] = z^\top \E \left[ \tilde{B}^\top \tilde{B} \right] z = \|z\|^2,
$$
where $\E \left[ \tilde{B}^\top \tilde{B} \right] = I$ is exploited. Let $z = P_{\perp_{W^{[t-1]}}} w^{(t)}$. Then, by applying the above equality to the expansion of $u^{(t+1)}$ in~\eqref{eqn:ut_expansion}, we have 
$$
\E \left[ \| u^{(t+1)} \|^2 \right] = \frac{d-t}{d} \cdot \|P_{\perp_{W^{[t-1]}}} w^{(t)}\|^2 
\in \left[ (\delta'_t)^2 \frac{k}{d^2} \left( 1-\frac{t}{d} \right), (\Delta'_t)^2 \frac{k}{d^2} \right], 
$$
where $\dim(X^{[t]}) = t$ is also used in the first step, and the last step is concluded from Hypothesis~\ref{hyp:proj_w}. Finally, by concentration property of random Gaussian vectors, when $t\ll d$ we have with high probability 
$$\|u^{(t+1)}\| \in \left[ \frac{\delta'_t}{2} \frac{\sqrt{k}}{d}, 2\Delta'_t \frac{\sqrt{k}}{d} \right].$$

\eprf

\subsubsection*{Hypothesis~\ref{hyp:proj_w}}

\paragraph{Computing $y^{(t)}$:}
In the first step of iteration $t$, the algorithm computes $y^{(t)}$. By induction Hypothesis~\ref{hyp:prog}, we know $|y_1^{(t)}| = \tilde{\Theta}(d^{\beta 2^{t-1}}\sqrt{k}/d)$. The other coordinates of $y^{(t)} := A^\top x^{(t)}$ are $y^{(t)}_{-1} = B^\top x^{(t)}$ which conditioning on the previous iterations are equivalent (in distribution) to 
\begin{align}
y^{(t)}_{-1} & \eqd \left( B^{(t-1,2)} \right)^\top x^{(t)} \nn \\
& = \left(\sum_{i \in [t-1]} \left( \frac{ u^{(i+1)} (P_{\perp_{W^{[i-1]}}} w^{(i)})^\top }{\|P_{\perp_{W^{[i-1]}}} w^{(i)}\|^2}
+ \frac{ P_{\perp_{X^{[i-1]}}} x^{(i)} (v^{(i)})^\top }{\|P_{\perp_{X^{[i-1]}}} x^{(i)}\|^2} \right) + B_{\res}^{(t-1,2)} \right)^\top x^{(t)} \nonumber \\
 & = \sum_{i\in[t-1]} \left( \tilde{\Theta} \left( \frac{d^2}{k} \right) P_{\perp_{W^{[i-1]}}} w^{(i)} \inner{u^{(i+1)}, x^{(t)}}+\tilde{\Theta}(1)v^{(i)} \inner{P_{\perp_{X^{[i-1]}}}x^{(i)},x^{(t)}} \right) + v^{(t)}, \label{eq:y}
\end{align}
where form of $B^{(t-1,2)}$ in~\eqref{eqn:B2} is used in the second equality. The bounds on the norms come from Hypotheses~\ref{hyp:proj_x}~and~\ref{hyp:proj_w}. The last term is by definition $v^{(t)} := (B^{(t-1,2)}_{\res})^\top x^{(t)}$. Note that differences in $\polylog$ factors in the (upper and lower) bounds in Hypotheses~\ref{hyp:proj_x}~and~\ref{hyp:proj_w} are represented by notation $\tilde{\Theta}(\cdot)$.

We will establish subsequently that if $k>d$, the terms involving $v^{(i)}$'s in the above expansion dominate, and the terms involving $P_{\perp_{W^{[i-1]}}} w^{(i)}$'s have norm of a smaller order; see Claim~\ref{clm:yinfinity}.

%

\paragraph{Computing $w^{(t)}$:}
In the next step of the algorithm at iteration $t$, $w^{(t)}$ is computed for which we now argue if the induction hypothesis holds up to iteration $t$, both lower and upper bounds at iteration $t$ as $\|P_{\perp_{W^{[t-1]}}} w^{(t)}\| \in [\delta'_{t}, \Delta'_{t}] \frac{\sqrt{k}}{d}$ (see induction Hypothesis~\ref{hyp:proj_w}) also hold.

\paragraph{Lower bound:}
For the lower bound, intuitively the {\em fresh} random vector $v^{(t)}$ should bring enough randomness into $w^{(t)}$. We formulate that in the following lemma.

\torestate{Lemma}{lem:freshrandomv}{
Suppose $R$ and $R'$ are two subspaces in $\R^k$ with dimension at most $t \le \frac{k}{16 (\log k)^2}$. Let $p \in \R^k$ be an arbitrary vector, $z \in \R^{k}$ be a uniformly random Gaussian vector in the space orthogonal to $R$, and finally $w = (p+z)*(p+z)$. Then with high probability, we have
\[ \|P_{\perp_{R'}} w\| \ge \frac{\E[\|z\|^2]}{40\sqrt{k}}.\]
}

Recall that $w^{(t)}:= y_{-1}^{(t)} * y_{-1}^{(t)}$, and $y_{-1}^{(t)}$ is expanded in \eqref{eq:y} as sum of an arbitrary vector and a random Gaussian vector.
Applying above lemma with $R = R'= \spn(W^{[t-1]})$, we have with high probability
\[
\|P_{\perp_{W^{[t-1]} }} w^{(t)}\|
\ge \frac{\E[\| v^{(t)} \|^2]}{40\sqrt{k}}
\ge \frac{\delta_t^2}{160} \sqrt{k}/d,
\]
where Hypothesis~\ref{hyp:normuv} gives lower bound $\| v^{(t)} \| \geq \delta_{t}/2\sqrt{k/d}$ (used in the second inequality). By choosing $\delta'_t = \delta_t^2/160$ the lower bound in Hypothesis~\ref{hyp:proj_w} is proved.

\paragraph{Upper bound:}
In order to prove the upper bounds in Hypothesis~\ref{hyp:proj_w}, we follow the sequence of arguments below: 

\begin{center}
\begin{tikzpicture}
  [
    scale=1.0,
    operator/.style={circle,minimum size=0.4cm,inner sep=0mm,draw=black},
  ]
  \node at ($(0,0)$) {$\textnormal{Claim~\ref{clm:yinfinity}: }\|y^{(t)}_{-1}\|_\infty \xRightarrow{(\cdot)^2} 
\|w^{(t)}\|_\infty \xRightarrow{\textnormal{Lemma~\ref{lem:projinftynormbound}}}
\|P_{\perp_{W^{[t-1]}}} w^{(t)}\|_\infty \Rightarrow
\|P_{\perp_{W^{[t-1]}}} w^{(t)}\|$};
  \draw (-6.5,0.6) -- (6.5,0.6) -- (6.5,-0.6) -- (-6.5,-0.6) -- (-6.5,0.6);
 \end{tikzpicture}
\end{center}

First we prove a bound on the infinity norm of $y^{(t)}_{-1}$:

\begin{claim}[Upper bound on $\|y^{(t)}_{-1}\|_\infty$]\label{clm:yinfinity} We have
$$\|y^{(t)}_{-1}\|_\infty \le \frac{t}{\delta_t} \frac{\log d}{\sqrt{d}} + (t-1) \left(\frac{\Delta'_{t-1}}{\delta'_{t-1}} \right)^2 \frac{1}{\sqrt{k}} = \tilde{O} \left( \frac{1}{\sqrt{d}} \right).$$
\end{claim}

\bprf
We exploit the induction hypothesis to bound the $\ell_\infty$ norm of all the terms in the expansion of $y^{(t)}_{-1}$ in \eqref{eq:y}.

For the terms involving $v^{(i)}$, since they are random Gaussian vectors with expected square norm at most $k/d$, by Lemma~\ref{lem:randomgaussianinfinitynorm} we know $\|v^{(i)}\|_\infty \le \frac{\log d}{\sqrt{d}}$ with high probability. In addition, for $v^{(i)}$, $i < t$, the coefficient is bounded as
\begin{equation} \label{eqn:coeffbound2}
\frac{\inner{P_{\perp_{X^{[i-1]}}}x^{(i)},x^{(t)}}}{\|P_{\perp_{X^{[i-1]}}}x^{(i)}\|^2} \le \frac{1}{\|P_{\perp_{X^{[i-1]}}}x^{(i)}\|} \leq \frac{1}{\delta_i},
\end{equation}
where the last step uses Hypothesis~\ref{hyp:proj_x}.
Therefore, the total contribution from terms involving $v^{(i)}$ in $\|y^{(t)}_{-1}\|_\infty$ is bounded by $\frac{t}{\delta_t} \frac{\log d}{\sqrt{d}}$.

For the terms involving $P_{\perp_{W^{[i-1]}}} w^{(i)}, i \in [t-1]$, we have from Hypothesis~\ref{hyp:proj_w} that the $\ell_\infty$ norm is bounded as
$\| P_{\perp_{W^{[i-1]}}} w^{(i)} \|_\infty \leq \Delta'_i \frac{1}{d}.$
In addition, the corresponding coefficient is bounded by
\begin{equation} \label{eqn:coeffbound1}
\frac{\inner{u^{(i+1)},x^{(t)}}}{\|P_{\perp_{W^{[i-1]}}} w^{(i)}\|^2}
\leq \frac{\|u^{(i+1)}\| \cdot \|x^{(t)}\|}{\|P_{\perp_{W^{[i-1]}}} w^{(i)}\|^2}
\le \frac{2\Delta'_i}{{\delta'_i}^2} \frac{d}{\sqrt{k}}.
\end{equation}
Again bounds in Hypotheses~\ref{hyp:proj_w}~and~\ref{hyp:normuv} are exploited in the last inequality.
Hence, the total contribution from terms involving $P_{\perp_{W^{[i-1]}}} w^{(i)}, i \in [t-1]$ in $\|y^{(t)}_{-1}\|_\infty$ is bounded by $(t-1) \left(\frac{\Delta'_{t-1}}{\delta'_{t-1}} \right)^2 \frac{1}{\sqrt{k}}$.

Combining the above bounds finishes the proof.
\eprf

Since $w^{(t)}:= y_{-1}^{(t)} * y_{-1}^{(t)}$, the above claim immediately implies that
\beq \|w^{(t)}\|_\infty \leq \tilde{O} \left( \frac{1}{d} \right). \label{eqn:wt_inftynorm} \eeq
Now we have the $\ell_\infty$ norm on $w$, however we need to bound the $\ell_\infty$ norm of the projected vector $P_{\perp_{W^{[t-1]}}} w^{(t)}$. Intuitively this is clear as the vectors in the space $W^{[t-1]}$ all have small $\ell_\infty$ as guaranteed by induction hypothesis. We formalize this intuition using the following lemma.

\torestate{Lemma}{lem:projinftynormbound}{
Suppose $R$ is a subspace in $\R^k$ of dimension $t'$, such that there is a basis $\{r_1,\dotsc,r_{t'}\}$ with $\|r_i\|_\infty \le \frac{\Delta}{\sqrt{k}}$ and $\|r_i\| = 1$. Let $p \in \R^k$ be an arbitrary vector, then
$$\|P_{\perp_{R}} p\|_\infty \le \|p\|_\infty + \|p\| \Delta\frac{\sqrt{t'}}{\sqrt{k}}.$$
}

Let $R = \spn(W^{[t-1]})$. Then the vectors $P_{\perp_{W^{[i-1]}}} w^{(i)}/\|P_{\perp_{W^{[i-1]}}} w^{(i)}\|$, $i\in[t-1]$ form a basis for subspace $R$, and we know from Hypothesis~\ref{hyp:proj_w} that the $\ell_\infty$ norm of each of these basis vectors is bounded by $\frac{\Delta}{\sqrt{k}}$ for $\Delta := \frac{\Delta'_{t-1}}{\delta'_{t-1}}$ which is of order $\polylog d$.
Applying above lemma, we have
$$\|P_{\perp_{W^{[t-1]}}} w^{(t)}\|_\infty \le \|w^{(t)}\|_\infty (1+ \Delta \sqrt{t-1}) \leq \frac{\Delta'_t}{d},$$
where the last inequality uses bound~\eqref{eqn:wt_inftynorm}, and appropriate choosing for $\Delta'_t$ which is of order $\polylog d$ and only depends on $t$ and $\log d$. This concludes the upper bound on the $\ell_\infty$ norm in Hypothesis~\ref{hyp:proj_w}. The upper bound on the $\ell_2$ norm is also immediately argued using this $\ell_\infty$ norm bound where an additional $\sqrt{k}$ factor shows up.

\subsubsection*{Hypothesis~\ref{hyp:proj_x}}
\paragraph{Computing $x^{(t+1)}$:}

In the next step of iteration $t$, the algorithm computes $x^{(t+1)}$. Conditioning on the previous iterations, the unnormalized version $\tilx^{(t+1)}$ is equivalent (in distribution) to

\begin{align}
\tilx^{(t+1)} & \eqd B^{(t,1)} w^{(t)} + (y_1^{(t)})^2 a_1 \nn \\
& = \sum_{i \in [t-1]} \frac{ u^{(i+1)} (P_{\perp_{W^{[i-1]}}} w^{(i)})^\top }{\|P_{\perp_{W^{[i-1]}}} w^{(i)}\|^2} w^{(t)} + \sum_{i\in [t]} \frac{ P_{\perp_{X^{[i-1]}}} x^{(i)} (v^{(i)})^\top }{\|P_{\perp_{X^{[i-1]}}} x^{(i)}\|^2} w^{(t)} + B_{\res}^{(t,1)} w^{(t)} + (y_1^{(t)})^2 a_1 \nn \\
 & = \sum_{i\in[t-1]} \tilde{\Theta} \left( \frac{d^2}{k} \right) u^{(i+1)} \inner{P_{\perp_{W^{[i-1]}}} w^{(i)} ,w^{(t)}}+\sum_{i\in[t]} \tilde{\Theta}(1)P_{\perp_{X^{[i-1]}}} x^{(i)}  \inner{ v^{(i)},w^{(t)}} + u^{(t+1)} + (y_1^{(t)})^2 a_1, \label{eq:x}
\end{align}
where form of $B^{(t,1)}$ in~\eqref{eqn:B1} is used in the second equality. The bounds on the norms come from Hypotheses~\ref{hyp:proj_x}~and~\ref{hyp:proj_w}. The last term is the definition of $u^{(t+1)} := B^{(t,1)}_{\res} w^{(t)}$. Note that differences in $\polylog$ factors in the (upper and lower) bounds in Hypotheses~\ref{hyp:proj_x}~and~\ref{hyp:proj_w} are represented by notation $\tilde{\Theta}(\cdot)$.

The goal is to prove Hypothesis~\ref{hyp:proj_x} holds at $t$-th iteration (which is to show the desired lower and upper bounds on $\|P_{\perp_{X^{[t]}}} x^{(t+1)} \|$) assuming induction hypothesis holds for earlier iterations. Given the normalization $x^{(t+1)} := \tilx^{(t+1)}/ \|\tilx^{(t+1)}\|$ in each iteration, we have
\beq \label{eqn:projx}
\|P_{\perp_{X^{[t]}}} x^{(t+1)} \| = \frac{1}{\|\tilx^{(t+1)}\|} \|P_{\perp_{X^{[t]}}} \tilx^{(t+1)} \|.
\eeq
Therefore, we first bound the norm of $\tilx^{(t+1)}$ which turns out to be $\|\tilx^{(t+1)}\| = \tilde{\Theta} \left( \frac{\sqrt{k}}{d} \right)$ as argued in the following.

\paragraph{Lower bound:}
The lower bound on $\|\tilx^{(t+1)}\|$ simply follows from the term $u^{(t+1)}$, which is an independent random Gaussian.

\begin{claim}
If $t \le \frac{d}{10}$, then we have whp
$$\|\tilx^{(t+1)}\| \ge \frac{\delta'_t}{4} \frac{\sqrt{k}}{d}.$$
\end{claim}

\bprf
We have
$$
\|\tilx^{(t+1)}\| \ge \|P_{\spn(X^{[t]}, U^{[t]}, a_1)^\perp}\tilx^{(t+1)}\|=\|P_{\spn(X^{[t]}, U^{[t]}, a_1)^\perp}u^{(t+1)}\|.
$$
Note that the equality is concluded from expansion of $\tilx^{(t+1)}$ in \eqref{eq:x} where all the components of $\tilx^{(t+1)}$ in the subspace $\spn(X^{[t]}, U^{[t]}, a_1)^\perp$ is represented by $u^{(t+1)}$.
The vector $P_{\spn(X^{[t]}, U^{[t]}, a_1)^\perp}u^{(t+1)}$ is the projection of a random Gaussian vector $u^{(t+1)}$ in to a subspace of dimention $d - o(d)$. Hence it is still a random Gaussian vector with expected square norm larger than $\frac{{\delta'_t}^2}{2} \frac{k}{d^2}$. 
By Lemma~\ref{lem:randomgaussiannorm}, with high probability the desired bound holds.
\eprf


\paragraph{Upper bound:} The upper bound is argued in the following claim.

\begin{claim}
We have either $$\inner{x^{(t+1)}, a_1} \ge 1-\gamma,$$ for some constant $\gamma>0$ or $$\|\tilx^{(t+1)}\| \le \tilde{O} \left( \frac{\sqrt{k}}{d} \right).$$
\end{claim}

\bprf
Let $\tilx^{(t+1)}$ in~\eqref{eq:x} be written as $\tilx^{(t+1)} = z+(y_1^{(t)})^2 a_1$ where vector $z \in \Rbb^d$ represents all the other terms in the expansion. The analysis is done under two cases 1) $(y_1^{(t)})^2 \geq \frac{2}{\gamma} \|z\|$ and 2) $(y_1^{(t)})^2 < \frac{2}{\gamma} \|z\|$ for some constant $\gamma>0$. Note that the left hand side is the norm of $(y_1^{(t)})^2 a_1$ since $\|a_1\|=1$, and in addition $(y_1^{(t)})^2 = \inner{x^{(t)},a_1}^2$.

{\em Case 1} $\left( (y_1^{(t)})^2 \geq \frac{2}{\gamma} \|z\| \right)$:
For the $x^{(t+1)} := \tilx^{(t+1)}/\|\tilx^{(t+1)}\|$, we have
\begin{align*}
\inner{x^{(t+1)},a_1} &= \frac{1}{\| z+(y_1^{(t)})^2 a_1 \|} \inner{z+(y_1^{(t)})^2 a_1,a_1} \\
& \geq \frac{1}{\| z\|+(y_1^{(t)})^2} \left[ (y_1^{(t)})^2 - \|z\| \right] \\
& \geq \frac{1-\frac{\gamma}{2}}{1+\frac{\gamma}{2}} \geq 1 - \gamma,
\end{align*}
where triangle and Cauchy-Schwartz inequality are used in the first bound, and the second inequality is concluded from assumption $(y_1^{(t)})^2 \geq \frac{2}{\gamma} \|z\|$.

{\em Case 2} $\left( (y_1^{(t)})^2 < \frac{2}{\gamma} \|z\| \right)$: We exploit the induction hypothesis to bound the norm of all the terms in the expansion of $\tilx^{(t+1)}$ in \eqref{eq:x}.

For the terms involving $u^{(i+1)}, i \in [t]$, we have $\|u^{(i+1)}\| \leq 2\Delta'_i \frac{\sqrt{k}}{d}$ from Hypothesis~\ref{hyp:normuv} and the argument for $\|u^{(t+1)}\|$. In addition, for $u^{(i+1)}$, $i \in [t-1]$, the coefficient is bounded as
\beq \frac{\inner{P_{\perp_{W^{[i-1]}}}w^{(i)},w^{(t)}}}{\|P_{\perp_{W^{[i-1]}}}w^{(i)}\|^2} \le \frac{\|w^{(t)}\|}{\|P_{\perp_{W^{[i-1]}}}w^{(i)}\|} \leq \frac{\Delta'_t}{\delta'_i}, \label{eqn:coeff_u} \eeq
where Cauchy-Schwartz inequality is used in the first bound, and the bound in Hypothesis~\ref{hyp:proj_w} and \eqref{eqn:wt_inftynorm} are exploited in the last inequality.
Therefore, the total contribution from terms involving $u^{(i+1)}$ in $\|\tilx^{(t+1)}\|$ is bounded by $\frac{2(t-1){\Delta'_t}^2}{\delta'_t} \frac{\sqrt{k}}{d}$.

For the terms involving $P_{\perp_{X^{[i-1]}}}x^{(i)}, i \in [t]$, we have $\|P_{\perp_{X^{[i-1]}}}x^{(i)}\| \leq 1$, but the coefficient $\inner{v^{(i)},w^{(t)}}$ needs further analysis to be bounded which is done in Lemma~\ref{lem:hardcorrbound} saying $|\inner{v^{(i)},w^{(t)}}| \leq \tilde{O} \left( \frac{\sqrt{k}}{d} \right)$. This implies that the total contribution from terms involving $P_{\perp_{X^{[i-1]}}}x^{(i)}$ in $\|\tilx^{(t+1)}\|$ is bounded by $\tilde{O} \left( \frac{\sqrt{k}}{d} \right)$.

Combining the above bounds and considering the assumption that the norm of $(y_1^{(t)})^2 a_1$ in the expansion of $\tilx^{(t+1)}$ is dominated by the norm of other terms argued above, the proof is complete concluding that $\| \tilx^{(t+1)} \| \leq \tilde{O} \left( \frac{\sqrt{k}}{d} \right)$.
\eprf

\torestate{Lemma}{lem:hardcorrbound}{
Under the induction hypothesis (up to update step $\tilx^{(t+1)} := A (y^{(t)})^{*2}$ at iteration $t$), we have for $i \in [t]$,
$$|\inner{v^{(i)},w^{(t)}}| \le O \left( t^3 \frac{(\Delta'_{t-1})^4}{(\delta'_{t-1})^4 \delta_t^2} (\log d) \frac{\sqrt{k}}{d} \right) = \tilde{O} \left( \frac{\sqrt{k}}{d} \right).$$
}

Using \eqref{eqn:projx} and the fact that $\|\tilx^{(t+1)}\| = \tilde{\Theta} \left( \frac{\sqrt{k}}{d} \right)$, we have
$$
\|P_{\perp_{X^{[t]}}} x^{(t+1)} \|
\geq \tilde{\Theta} \left( \frac{d}{\sqrt{k}} \right) \|P_{\spn(X^{[t]}, U^{[t]}, a_1)^\perp}u^{(t+1)}\| \geq \frac{\delta'_t}{4},
$$
where the bound $\|P_{\spn(X^{[t]}, U^{[t]}, a_1)^\perp}u^{(t+1)}\| \ge \frac{\delta'_t}{4} \frac{\sqrt{k}}{d}$ is also used. This finishes the proof that Hypothesis~\ref{hyp:proj_x} holds. 

\subsubsection*{Hypothesis~\ref{hyp:prog}}

Finally we prove Hypothesis~\ref{hyp:prog} at iteration $t$ given earlier induction hypothesis. The first part of the hypothesis is proved in the following claim.

\begin{claim} \label{clm:corr}
We have
$$|\inner{a_1,x^{(t+1)}}| \in [\delta^*_{t+1}, \Delta^*_{t+1}] d^{\beta 2^t} \frac{\sqrt{k}}{d}.$$
\end{claim}
\bprf
We first show the correlation bound on the unnormalized version as $\inner{a_1,\tilx^{(t+1)}}$.
Looking at the expansion of $\tilx^{(t+1)}$ in~\eqref{eq:x}, the correlation $\inner{a_1,\tilx^{(t+1)}}$ involves three types of terms emerging from $(y_1^{(t)})^2 a_1$, $u^{(i+1)}$ and $P_{\perp_{X^{(i-1)}}}x^{(i)}$. In the following, we argue the correlation from each of these terms where we observe that the correlation is dominated by the term $(y_1^{(t)})^2 a_1$, and the rest of terms contribute much smaller amount.

For the term $(y_1^{(t)})^2 a_1$, we have
$$
\inner{a_1,(y_1^{(t)})^2 a_1} = (y_1^{(t)})^2 \in [(\delta_t^*)^2,(\Delta_t^*)^2] d^{\beta 2^t} \frac{k}{d^2},
$$
where the last part exploits induction Hypothesis~\ref{hyp:prog} in the previous iteration.

For the terms involving $u^{(i+1)}$, these vectors are random Gaussian vectors in a subspace (with dimension $\Omega(d)$), and therefore, we have with high probability
$$\inner{a_1,u^{(i+1)}} \leq \E[\|u^{(i+1)}\|] \cdot O \left( \frac{\log d}{\sqrt{d}} \right) \leq \tilde{O} \left( \frac{\sqrt{k}}{d\sqrt{d}} \right) \le \tilde{O} \left( \frac{k}{d^2} \right),$$
where the correlation bound between two independent random Gaussian vectors in $\Omega(d)$-dimension is used in the first inequality\,\footnote{For two independent random Gaussian vectors $p,q \in \Rbb^d$, we have with high probability $\inner{p,q} \le \E[\|p\|] \cdot \E[\|q\|] \cdot O \left( \frac{\log d}{\sqrt{d}} \right)$.}, Hypothesis~\ref{hyp:normuv} is exploited in the second inequality, and finally last inequality is from assumption $k>d$. In addition, the coefficient associated with $u^{(i+1)}$ is bounded by $\Delta'_t/\delta'_i$ argued in~\eqref{eqn:coeff_u}. Hence, the total contribution from terms involving $u^{(i+1)}$ in $\inner{\tilx^{(t+1)},a_1}$ is bounded by $\tilde{O} \left( \frac{k}{d^2} \right)$.

For the terms involving $P_{\perp_{X^{[i-1]}}}x^{(i)}$, by Hypothesis~\ref{hyp:prog} we have
$$
\inner{a_1,P_{\perp_{X^{[i-1]}}}x^{(i)}} \leq \Delta^*_i d^{\beta 2^{i-1}} \frac{\sqrt{k}}{d}.
$$
In addition, the associated coefficient is bounded by $\tilde{O} \left( \frac{\sqrt{k}}{d} \right)$ from Lemma~\ref{lem:hardcorrbound}. Hence, the total contribution from terms involving $P_{\perp_{X^{[i-1]}}}x^{(i)}$ in $\inner{\tilx^{(t+1)},a_1}$ is bounded by $\tilde{O} \left( d^{\beta 2^{t-1}} \frac{k}{d^2} \right)$.

Combining the above bounds implies
$$|\inner{a_1,\tilx^{(t+1)}}| \leq \tilde{O} \left( d^{\beta 2^t} \frac{k}{d^2} \right).$$
Finally, using the bound on the norm of $\tilx^{(t+1)}$ argued as $\|\tilx^{(t+1)}\| = \tilde{\Theta} \left( \frac{\sqrt{k}}{d} \right)$ finishes the proof.
\eprf

To prove the last part of Hypothesis~\ref{hyp:prog}, we use the following lemma which is very similar to Lemma~\ref{lem:projinftynormbound}.

\torestate{Lemma}{lem:projinnerproduct}{
Suppose $R$ is a subspace in $\R^d$ of dimension $t'$, such that there is a basis $\{r_1,\dotsc,r_{t'}\}$ with $|\inner{r_i,a_1}| \le \Delta$ and $\|r_i\| = 1$. Let $p \in \R^d$ be an arbitrary vector, then
$$|\inner{P_{\perp_{R}} p,a_1}| \le |\inner{p,a_1}| + \|p\| \Delta \sqrt{t'}.$$
}
We apply this lemma with $R = \spn(X^{[t]})$, and the basis is $P_{\perp X^{[i-1]}}X^{(i)}/\|P_{\perp X^{[i-1]}}X^{(i)}\|$. By induction hypothesis $\Delta$ in the lemma is at most $\Delta_*^t d^{\beta 2^t}\sqrt{k}/d$, let $v = x^{(t+1)}$ then this gives the desired bound.

Let $R = \spn(X^{[t]})$. Then the vectors $P_{\perp_{X^{[i-1]}}}x^{(i)}/\|P_{\perp_{X^{[i-1]}}}x^{(i)}\|$, $i \in [t]$ form a basis for subspace $R$, and we know from Hypotheses~\ref{hyp:proj_x}~and~\ref{hyp:prog} that the correlation between these basis vectors and $a_1$ is bounded by $\Delta := \Delta^*_t d^{\beta 2^{t-1}}\frac{\sqrt{k}}{d}$.
Applying above lemma, we have
$$|\inner{P_{\perp_{X^{[t]}}} x^{(t+1)},a_1}| \le |\inner{x^{(t+1)},a_1}| + \Delta \sqrt{t} \leq \Delta^*_{t+1} d^{\beta 2^t} \frac{\sqrt{k}}{d},$$
where the last inequality uses the first part of Hypothesis~\ref{hyp:prog} proved earlier in this section. Note that $\Delta^*_{t+1}$ is a new $\polylog$ factor here.

\subsection{Growth rate of  $\delta_t$, $\delta'_t$, $\Delta'_t$, $\delta^*_t$, $\Delta^*_t$} \label{appendix:delta}

We know that if the number of iterations $t$ is a constant, then the $\delta$ and $\Delta$ parameters (i.e., $\delta_t$, $\delta'_t$, $\Delta'_t$, $\delta^*_t$, $\Delta^*_t$) in the induction hypothesis are bounded by polylog factors of $d$. Here, we show that these parameters can be still bounded even when the number of steps is slightly larger than a constant.
Let $$R_t := \max\{1/\delta_t, \Delta'_{t-1}/\delta'_{t-1}, \Delta_t^*/\delta_t^*\}.$$ We know $R_1 = 1$, and by the inductive step analysis we have the following polynomial recursion property.

\begin{claim}
$R_{t+1} = \poly(R_t, t, \log d)$.
\end{claim}

This claim follows from the proof of inductive step, where in every step the $\delta$ and $\Delta$ parameters are bounded by polynomial functions of previous $\delta$'s ($\Delta$'s), $t$, and $\log d$.

We now solve this recursion as follows.

\begin{lemma}
\label{lem:recursion}
Suppose $R_{t+1} \le c_0 R_t^{c_1} t^{c_2} (\log d)^{c_3}$ where $c_0,c_1,c_2,c_3$ are positive constants, and we know $R_1 = 1$. Then 
$$R_t \le (\log d)^{2^{c_4 t}},$$ 
for some constant $c_4>0$ depending on $c_0,c_1,c_2,c_3$.
\end{lemma}

\bprf
Without loss of generality assume $c_0 \ge 1$, $c_2\ge 1$, $c_3\ge 1$, and $R_1 \ge \log d$. Given these assumptions, we have $R_t \ge \max\{c_0, t,\log d\}$, for $t \ge 1$. Applying this to the assumption $R_{t+1} \le c_0 R_t^{c_1} t^{c_2} (\log d)^{c_3}$, we have
\begin{equation} \label{eqn:recursionaux}
R_{t+1} \le R_t^{1+c_1+c_2+c_3}.
\end{equation}
Pick some $q>0$ such that $R_1 \le  (\log d)^{2^{q}}$, and pick some $$c_4 \ge \max\{q, \log_2(1+c_1+c_2+c_3)\}.$$ 

Now we prove the result by the induction argument. Since $c_4 \ge q$, the basis of induction holds for $R_1$.
As the inductive step, suppose $R_t\le (\log d)^{2^{c_4 t}}$. Applying this to \eqref{eqn:recursionaux}, we have
$$R_{t+1} \le (\log d)^{(1+c_1+c_2+c_3)2^{c_4t}} \leq (\log d)^{2^{c_4(t+1)}},$$
where $2^{c_4} \ge (1+c_1+c_2+c_3)$ is used in the last inequality. This finishes the inductive step and the result is proved.
\eprf

Using the above bound, we show in the following corollary that the $\delta$ and $\Delta$ parameters in the induction hypothesis are bounded by polylog factors of $d$ even if the number of steps $t$ goes up to $c \log \log d$ for small enough constant $c$. 
In addition, we show that if $\beta \ge (\log d)^{-c_5}$ for some constant $c_5>0$, then the power method 
converges to a point $x^{(t)}$ which is constant close to the true component.

\begin{corollary} \label{corr:betabound}
There exists a universal constant $c_5 > 0$ such that if $$\beta \ge (\log d)^{-c_5},$$ and the initial correlation is lower bounded by $d^{\beta} \frac{\sqrt{k}}{d}$ (see~\eqref{eqn:init assumption}), then with high probability the power method gets to a point that is constant close  to the true component in $\Theta(\log \log d)$ number of steps.
\end{corollary}

\bprf
Pick  the number of steps to be $t = (\log \log d)/2c_4$, where $c_4$ is the constant in Lemma~\ref{lem:recursion}. Then, from Lemma~\ref{lem:recursion} we have 
$$R_t \le (\log d)^{\sqrt{\log d}} \le o(d),$$ 
where the last inequality can be shown by taking the log of both sides. This says that the analysis of inductive step still holds after such number of iterations.

Finally, by progress bound in~\eqref{eqn:progress}, we can see that if $\beta \ge (\log d)^{-c_5}$, then the power method converges to a point $x^{(t)}$ which is constant close to the true component.
\eprf

\section{Auxiliary Lemmas for Induction Argument}

In this section we prove the lemmas used in arguing inductive step in Appendix~\ref{sec:inductivestep}. 

We first introduce the following lemma proposing a lower bound on the singular value of  product of matrices.

\begin{lemma}[\citealt{MerikoskiKumar2004}] \label{lem:Lidskii'sInequality}
Let $C$ and $D$ be $k \times k$ matrices. If $1 \le i \le k$ and $1 \le l \le k-i+1$, then
$$\sigma_{i}(CD) \geq \sigma_{i+l-1}(C) \cdot \sigma_{k-l+1}(D),$$
where $\sigma_j(C)$ denotes the $j$-th singular value (in decreasing order) of matrix $C$.
\end{lemma}

\subsection{Properties of random Gaussian vectors}
We start with some basic properties of random Gaussian vectors. First as a simple fact, the norm of a random Gaussian vector is concentrated as follows which is proved via simple concentration inequalities.

\begin{lemma}\label{lem:randomgaussiannorm}
Let $z \in \R^d$ be a random Gaussian vector with $\E[zz^\top] = \frac{1}{d} I$. Then we have with high probability $\frac{1}{2} \le \|z\| \le 2$.
\end{lemma}

Next we show the $\ell_\infty$ norm of a Gaussian vector is small, even if it is projected on some subspace.

\begin{lemma}\label{lem:randomgaussianinfinitynorm}
Let $R$ be any linear subspace in $\R^d$ and $z\in \R^d$ be a random Gaussian vector with $\E[zz^T] = \frac{1}{d} I$. Then we have with high probability $\|P_{\perp_R} z\|_\infty \le  \frac{\log d}{\sqrt{d}}$.
\end{lemma}

\bprf
Since $P_{\perp_R}$ is a projection matrix, in particular the norm of its columns is bounded by $1$. Hence, each entry of $P_{\perp_R}z$ is a Gaussian random variable with variance bounded by $\frac{1}{d}$ implying that with high probability the absolute value of each coordinate is smaller than $\frac{\log d}{\sqrt{d}}$. Finally, the desired $\ell_\infty$ norm bound is argued by applying union bound.
\eprf

We can also show that most of the entries are of size at least $\frac{1}{\sqrt{d}}$.

\begin{lemma}\label{lem:randomgaussianlower}
Let $R$ be any linear subspace in $\R^d$ with dimension $t \le \frac{d}{16 (\log d)^2}$ and $z\in \R^d$ be a random Gaussian vector with $\E[zz^T] = \frac{1}{d} I$. Then we have with high probability at least $1/2$ of the entries $i\in[d]$ satisfy $|(P_{\perp_R} z)_i| \ge \frac{1}{4\sqrt{d}}$.
\end{lemma}

\bprf
Since the entries of $z$ are independent Gaussian random variables with standard deviation $\frac{1}{\sqrt{d}}$, we know with high probability at least $1/2$ of the entries have absolute value larger than $\frac{1}{2\sqrt{d}}$. On the other hand, $P_R z$ is also a random Gaussian vector with expected square norm bounded by
$$\E[\|P_R z \|^2] \leq \frac{\E[\|z\|^2]}{16 (\log d)^2} = \frac{1}{16 (\log d)^2},$$
where the assumption on the dimension of subspace $R$ is used in the inequality. By Lemma~\ref{lem:randomgaussianinfinitynorm} we know with high probability entries of $P_R z$ are bounded by $1/4\sqrt{d}$. Now $P_{\perp_R} z = z - P_R z$ must have at least $1/2$ of the entries with absolute value larger than $1/4\sqrt{d}$.
\eprf

Using the above lemmas, we can prove Lemma~\ref{lem:freshrandomv}.

\restate{lem:freshrandomv}

\bprf
Let $z, z'$ be two independent samples of $z$, and $w,w'$ be the corresponding $w$ vectors. We have 
\begin{equation} \label{eqn:w_expansion}
w - w' = (p+z)*(p+z)-(p+z')*(p+z') = (2p+z+z')*(z-z').
\end{equation}
By properties of Gaussian vectors, $z+z'$, $z-z'$ are two {\em independent} random Gaussian vectors in the subspace orthogonal to $R$ each with expected square norm $2\E[\|z\|^2]$. We use $z_1 := z+z'$ and $z_2 := z-z'$ to denote these two random Gaussian vectors.

Next, we show that with high probability 
$$\|P_{\perp_{R'}} (w - w')\| \ge \frac{\E[\|z\|^2]}{20\sqrt{k}}.$$ Note that this implies the result of lemma as follows. Suppose $\|P_{\perp_{R'}} w\| < \frac{1}{40} \E[\|z\|^2]/\sqrt{k}$ with probability $\delta$. Since $w'$ is an independent sample, with probability $\delta^2$ this bound holds for both $w$ and $w'$. When this happens, we have $\|P_{\perp_{R'}} (w - w')\| < \frac{1}{20} \E[\|z\|^2]/\sqrt{k}$ by triangle inequality. Since we showed $\delta^2$ is negligible, $\delta$ is also negligible.

First we sample $z_2$. Let $R'' = \spn(R', p*z_2)$. Then by expansion of $w-w'$ in~\eqref{eqn:w_expansion}, we have
\begin{equation} \label{eqn:aux_freshrandomv}
\|P_{\perp_{R'}}(w-w')\| = \|P_{\perp_{R'}} \bigl( 2(p*z_2) + (z_1*z_2) \bigr)\| \ge \|P_{\perp_{R''}} (z_1*z_2)\| = \|P_{\perp_{R''}}\mbox{Diag}(z_2)P_{\perp_R} z_1\|,
\end{equation}
where the inequality is concluded by ignoring the component along $p*z_2$  direction. The last equality is from\,\footnote{For vector $u$, $\Diag(u)$ denotes the diagonal matrix with $u$ as its main diagonal.} $u*v = \Diag(u) \cdot v$ (for two vectors $u$ and $v$), and the assumption that $z_1 = z+z'$ is in the subspace orthogonal to $R$.
For the matrix $P_{\perp_{R''}}\mbox{Diag}(z_2)P_{\perp_R}$, we have\,\footnote{Recall that $\sigma_l(A)$ denotes the $l$-th singular value (in decreasing order) of matrix $A$.}
$$\sigma_{k/4} \left( P_{\perp_{R''}}\mbox{Diag}(z_2)P_{\perp_R} \right) 
\geq \sigma_{k/2} \left( \mbox{Diag}(z_2) \right) \cdot \sigma_{7k/8} \left( P_{\perp_R} \right) \cdot \sigma_{7k/8} \left( P_{\perp_{R''}} \right) 
\ge \frac{\sqrt{\E[\|z\|^2]}}{4\sqrt{k}},$$
where the first inequality is from Lemma~\ref{lem:Lidskii'sInequality}, and the last step is argued as follows. By Lemma~\ref{lem:randomgaussiannorm}, with high probability $z_2$ has square norm at least $\E[\|z_2\|^2]/2 = \E[\|z\|^2]$, and therefore, by Lemma~\ref{lem:randomgaussianlower} at least $k/2$ of its entries have absolute value larger than $\frac{1}{4} \sqrt{\E[\|z\|^2]}/\sqrt{k}$.
Therefore, we can restrict attention to the space spanned by the $k/4$ top singular vectors. In addition, within this subspace we have with high probability $\|z_1\|^2 \ge \E[\|z\|^2]/8$, and hence, 
$$\|P_{\perp_{R''}}\mbox{Diag}(z_2)P_{\perp_R} z_1\| \ge \frac{\E[\|z\|^2]}{20\sqrt{k}},$$
which finishes the proof by applying ~\eqref{eqn:aux_freshrandomv}.
\eprf

\subsection{Properties of projections}

In this part we prove some basic properties of projections. Intuitively, if the whole subspace has small inner-product with some vector, then the projection of an arbitrary vector to the orthogonal subspace should not change the inner-product with that particular vector by too much. This is what we require in Lemma~\ref{lem:projinnerproduct}.

\restate{lem:projinnerproduct}

\bprf
We have $P_{\perp_R}p = p - \sum_{i=1}^{t'} \inner{p,r_i} r_i$, and therefore
\begin{align*}
|\inner{P_{\perp_R}p, a_1}| & \le |\inner{p,a_1}| + \sum_{i=1}^{t'} |\inner{p,r_i}\inner{a_1,r_i}|\\
& \le |\inner{p,a_1}| + \Delta \sum_{i=1}^{t'} |\inner{p,r_i}|\\
& \le |\inner{p,a_1}| + \Delta\sqrt{t'\sum_{i=1}^{t'}\inner{p,r_i}^2} \\
& \le |\inner{p,a_1}| + \Delta\|p\|\sqrt{t'}.
\end{align*}
The first step is triangle inequality and the third is Cauchy-Schwartz.
\eprf

Lemma~\ref{lem:projinftynormbound} is very similar.

\restate{lem:projinftynormbound}

This lemma essentially follows from Lemma~\ref{lem:projinnerproduct}, because $\ell_\infty$ norm is just the maximum inner-product to a basis vector. More specifically, the above lemma is applied for all $a_1 = e_j, j \in [k]$, where $e_j$ denotes the $j$-th basis vector in $\R^k$.

\subsection{Bounding correlation between $v$ and $w$}

We are only left with Lemma~\ref{lem:hardcorrbound}. The main difficulty in proving this lemma is that the later steps are dependent on the previous steps. In the proof we show the dependency is bounded and in fact we can treat them as independent.

\restate{lem:hardcorrbound}

\bprf
Recall $w^{(t)} =  y^{(t)}_{-1}*y^{(t)}_{-1}$, and $y^{(t)}_{-1}$ is specified in \eqref{eq:y}. We now expand the Hadamard product in $w^{(t)}$ and bound all the resulting $O(t^2)$ terms. 

The first type of terms has the form $\inner{v^{(i)}, P_{\perp_{W^{[i_1-1]}}} w^{(i_1)}*P_{\perp_{W^{[i_2-1]}}} w^{(i_2)}}$, which can be bounded as
\begin{align*}
\inner{v^{(i)}, P_{\perp_{W^{[i_1-1]}}} w^{(i_1)}*P_{\perp_{W^{[i_2-1]}}} w^{(i_2)}} 
& \leq k\cdot \|v^{(i)}\|_\infty \cdot \|P_{\perp_{W^{[i_1-1]}}} w^{(i_1)}*P_{\perp_{W^{[i_2-1]}}} w^{(i_2)}\|_\infty \\
& \le 2 k \frac{\log d}{\sqrt{d}}  \frac{(\Delta'_{t-1})^2}{d^2},
\end{align*}
where $\|v^{(i)}\|_\infty$ is bounded by Lemma~\ref{lem:randomgaussianinfinitynorm}, and $\ell_\infty$ norm of other vector is bounded by induction Hypothesis~\ref{hyp:proj_w}.
In addition, the corresponding coefficient is bounded by (see \eqref{eqn:coeffbound1}, and note that both $i_1,i_2 < t$)
$$\frac{4(\Delta'_{t-1})^2}{(\delta'_{t-1})^4} \frac{d^2}{k}.$$
Hence, the total contribution from such terms is bounded by 
\sublabon{equation}
\begin{equation} \label{eqn:aux1}
8 t^2 \frac{(\Delta'_{t-1})^4}{(\delta'_{t-1})^4} \frac{\log d}{\sqrt{d}}.
\end{equation}

The second type of terms has the form $\inner{v^{(i)}, P_{\perp_{W^{[i_1-1]}}} w^{(i_1)}*v^{(i_2)}} = \inner{v^{(i)}*v^{(i_2)}, P_{\perp_{W^{[i_1-1]}}} w^{(i_1)}}$, which can be bounded as
$$\|P_{\perp_{W^{[i_1-1]}}} w^{(i_1)}\|_\infty \cdot \|v^{(i)}*v^{(i_2)}\|_1 \le \|P_{\perp_{W^{[i_1-1]}}} w^{(i_1)}\|_\infty \cdot \frac{\|v^{(i)}\|^2+ \|v^{(i_2)}\|^2}{2} \le 4 \Delta_{t-1}' \frac{k}{d^2},$$
where the last inequality is concluded from Hypotheses~\ref{hyp:proj_w}~and~\ref{hyp:normuv}.
In addition, the corresponding coefficient is bounded by (see \eqref{eqn:coeffbound2} and \eqref{eqn:coeffbound1}, and note that both $i_1,i_2 < t$)
$$\frac{2\Delta'_{t-1}}{(\delta'_{t-1})^2\delta_{t-1}} \frac{d}{\sqrt{k}}.$$
Hence, the total contribution from such terms is bounded by 
\begin{equation} \label{eqn:aux2}
8t^2 \frac{(\Delta'_{t-1})^2}{\delta_{t-1}(\delta'_{t-1})^2} \frac{\sqrt{k}}{d}.
\end{equation}

The third type of terms has the form $\inner{v^{(i)}, v^{(i_1)}*v^{(i_2)}}$, with coefficient bounded by $1/\delta_{t-1}^2$ (see \eqref{eqn:coeffbound2}). For bounding these inner products, we need to use the fact that they are random Gaussian vectors, however the main difficulty is that they are correlated (if $i>j$, then the subspace that $v^{(i)}$ is in that depends on $v^{(j)}$). To resolve this difficulty, we treat $v^{(i)} \in \R^{k-1}$ as projection of $n^{(i)} \in \R^{k-1}$ into subspace orthogonal to $W^{[t-1]}$, where $n^{(i)}$'s are {\em independent} Gaussian vectors in the full $k-1$ dimensional space.  Independent of the ordering of $i,i_1,i_2$, we have with high probability
$$\inner{n^{(i)},n^{(i_1)}*n^{(i_2)}} \le O \left( \frac{\sqrt{k}}{d\sqrt{d}} \right),$$  
since it is a sum of $k-1$ independent mean-0 entries each with variance $\frac{1}{d^3}$. 
On the other hand, from Hypothesis~\ref{hyp:normuv}, we have $\E[\|v^{(i)}\|^2] \le 4\frac{k}{d}$, and since vector $n^{(i)}-v^{(i)}$ is in the subspace $W^{[t-1]}$ with dimension $t$, we have
$$\E[\|n^{(i)}-v^{(i)}\|^2] \le O \left( \frac{t}{k} \right) \cdot \frac{4k}{d} = O \left( \frac{t}{d} \right),$$
and therefore, we have with high probability $\|n^{(i)}-v^{(i)}\| \le O(\sqrt{t/d})$ for all $i \in [t-1]$. 
Using this, the difference between $\inner{n^{(i)},n^{(i_1)}*n^{(i_2)}}$ and $\inner{v^{(i)}, v^{(i_1)}*v^{(i_2)}}$ can be bounded as
$$|\inner{n^{(i)},n^{(i_1)}*n^{(i_2)}} - \inner{v^{(i)}, v^{(i_1)}*v^{(i_2)}}|
\leq O \left( (\log k) t \frac{\sqrt{k}}{d\sqrt{d}} \right),$$
where the right hand side is the bound on the dominant term in the expansion of difference as
\begin{align*}
|\inner{n^{(i)},(n^{(i_1)}-v^{(i_1)})*(n^{(i_2)}-v^{(i_2)})}| 
& \leq \|n^{(i)}\| \cdot \| (n^{(i_1)}-v^{(i_1)})*(n^{(i_2)}-v^{(i_2)}) \| \\
& \leq O \left( (\log k) \sqrt{\frac{k}{d}} \right) \cdot O \left( \frac{t}{d} \right) \\
& = O \left( (\log k) t \frac{\sqrt{k}}{d\sqrt{d}} \right). 
\end{align*}
Here, the first inequality is the Cauchy-Schwartz, and the second inequality is from bound on the norm of random Gaussian vector $n^{(i)}$, and the bound on the norm of difference vectors $n^{(i_1)}-v^{(i_1)}$ stated earlier.
Hence, the total contribution from such terms is bounded by
\begin{equation} \label{eqn:aux3}
O \left( t^3 \frac{\log k}{\delta_{t-1}^2} \frac{\sqrt{k}}{d\sqrt{d}} \right).
\end{equation}
\sublaboff{equation}

Taking the sum of all the terms in \eqref{eqn:aux1}-\eqref{eqn:aux3} gives the desired bound.

\eprf

%

\section{Additional Arguments for Noise Analysis} \label{appendix:noise}

\bprfof{Lemma~\ref{lem:noise-induction}}
We prove this by an induction argument. 

\paragraph{Basis of induction:}
For $t=1$, $x^{(1)}$ is the initialization vector and thus, $\xi^{(1)} = 0$. Hence, the proposed bound holds for the basis of induction $t=1$.

\paragraph{Inductive step:}
Assuming the inductive hypothesis holds for step $t$, we prove it also holds for step $t+1$. We have 
\begin{align}
x^{(t+1)} + \xi^{(t+1)} & = \normalize \left( \hat{T}(x^{(t)} + \xi^{(t)},x^{(t)} + \xi^{(t)},I) \right) \nn \\
&= \normalize \left( T(x^{(t)},x^{(t)},I) + 2T(x^{(t)}, \xi^{(t)}, I) + T(\xi^{(t)},\xi^{(t)},I) + E (\hat{x}^{(t)},\hat{x}^{(t)},I) \right). \label{eqn:update-noisy}
\end{align}
The first term $T(x^{(t)},x^{(t)},I)$ corresponds to the main signal; recall that $x^{(t+1)} = \normalize(T(x^{(t)},x^{(t)},I))$ in the noiseless setting, where the unnormalized version $\tl{x}^{(t+1)} :=T(x^{(t)},x^{(t)},I)$ has norm at least $\tl{\Omega}(\sqrt{k}/d)$ which is argued in the induction argument for Hypothesis~\ref{hyp:proj_x}.
We now bound the desired property of noise terms in the above expansion.

For the second term, we break it into two terms as
$$2T(x^{(t)}, \xi^{(t)}, I) = 2\inner{x^{(t)},a_1}\inner{\xi^{(t)},a_1}a_1 + 2T'(x^{(t)}, \xi^{(t)}, I) =: p + q,$$
where $T' := \sum_{j>1} a_j \otimes a_j \otimes a_j$. Here $p := 2\inner{x^{(t)},a_1}\inner{\xi^{(t)},a_1}a_1$ corresponds to the multilinear form from first component of $T$, and $q:= 2T'(x^{(t)}, \xi^{(t)}, I)$ corresponds to the multilinear form from the rest of components.

For $q$, we apply Lemma~\ref{lem:mixednorm}. Note that since $\|x^{(t)}\|_{B^*} \le \tilde{O}(1/\sqrt{d})$, we get an extra $1/\sqrt{d}$ factor in the bound provided by Lemma~\ref{lem:mixednorm}, and therefore we have 
$$\|q\|_2 \le \tilde{O}(\epsilon d^{\beta 2^{t-1}}\sqrt{k}/d),$$
where we also used the induction hypothesis $\|\xi^{(t)}\| \le \tilde{O}(\epsilon d^{\beta2^{t-1}})$.

For $p$, we have
$$\|p\| = 2|\inner{x^{(t)},a_1}|\cdot|\inner{\xi^{(t)},a_1}| \leq \tilde{O} \left( \epsilon d^{\beta2^t} \sqrt{k}/d \right),$$
where the inequality is from the signal and noise induction hypotheses; see Equation~\eqref{eqn:progress} for the signal induction hypothesis.

The third term $T(\xi^{(t)},\xi^{(t)},I)$ has $\ell_2$ norm bounded as 
$$\|T(\xi^{(t)},\xi^{(t)},I)\| \leq \|T\| \|\xi^{(t)}\|^2 \le \tilde{O}(d^{\beta 2^{t}}\epsilon^2) \le \tilde{O}(\epsilon d^{\beta 2^{t}} \sqrt{k}/d),$$
where the first inequality uses the sub-multiplicative property, and the second inequality exploits the bounded norm of random tensor $T$ as $\|T\| \leq O(1)$, and the induction hypothesis in $t$-th step. The final inequality uses the assumption $\epsilon < o(\sqrt{k}/d)$ in the lemma.

The fourth term $E (\hat{x}^{(t)},\hat{x}^{(t)},I)$ has $\ell_2$ norm bounded by
$$\| E (\hat{x}^{(t)},\hat{x}^{(t)},I) \| \leq \|E\| \|\hat{x}^{(t)}\|^2 \leq \epsilon \sqrt{k}/d,$$
where we use the sub-multiplicative property in the first inequality, and the assumption on the norm of error tensor $E$ in the lemma, and the fact that $\|\hat{x}^{(t)}\|=1$ are exploited in the second inequality. 

Summarizing the above calculations on different terms of the update in~\eqref{eqn:update-noisy}, the signal plus noise vector before normalization is 
$$T(x^{(t)},x^{(t)},I) + 2T(x^{(t)}, \xi^{(t)}, I) + T(\xi^{(t)},\xi^{(t)},I) + E (\hat{x}^{(t)},\hat{x}^{(t)},I) =: \alpha x^{(t+1)} + z,$$
where $\alpha$ is a coefficient which is lower bounded as $\alpha \ge \tilde{\Omega}(\sqrt{k}/d)$. The vector $z$ also satisfies
\begin{equation} \label{eqn:z-bounds}
\|z\| \le \tilde{O}(\epsilon d^{\beta 2^t}\sqrt{k}/d),
\end{equation}
which is derived by combining the bounds we argued on the second, third and fourth terms.

Note that until the very last step we always have  $d^{\beta 2^{t}} \le o(d/\sqrt{k})$ (otherwise we are constantly close to the true component, and we are done).  In this case the norm of $z$ is negligible compared to $\alpha$ since $\|z\| \leq o(\alpha)$, and thus, the normalization factor is equal to $\|\alpha x^{(t+1)} + z\| = \alpha(1\pm o(1))$. Therefore, after the normalization, we have the noise vector $\xi^{(t+1)} = \alpha' x^{(t+1)} + \beta z$, where $|\alpha'| \leq  \|z\|/\alpha \leq o(1)$   
 and $|\beta| \le 2/\alpha \le \tilde{O}(d/\sqrt{k})$, hence we know $\|\xi^{(t+1)}\| \le \tilde{O}(\epsilon d^{\beta 2^t})$.

For the last step of the induction, the norm of $T(x^{(t)},x^{(t)},I)$ is also larger (it has norm $d^{\beta 2^t} k/d^2$, which is larger than $\sqrt{k}/d$ for the last step). Since $\epsilon < o(\sqrt{k}/d)$ we still know the noise is negligible.

\eprfof

\paragraph{Lemma on the property of $\|\cdot\|_*$ norm defined in Definition~\ref{def:starnorm}:}

\begin{lemma} \label{lem:mixednorm}
Consider a random tensor $T = \sum_{j \in [k]} a_j \otimes a_j \otimes a_j$ where $a_j$'s are zero-mean random Gaussian with expected unit norm. Let $A\in \R^{d\times k}$ be the matrix $[a_1,\dotsc,a_k]$, $T' = \sum_{j>1} a_j\otimes a_j\otimes a_j$ and $B\in \R^{d\times (k-1)}$ be the matrix $[a_2,a_3,\dotsc,a_k]$. Then for any vectors $u,v$ such that $\|u\|_{B^*} \le 1$ and $\|v\|_2 \le 1$, with high probability we have
$$
\|T'(u,v,I)\|_2 \le \tilde{O} \left( \sqrt{k/d} \right).
$$
\end{lemma}

\bprf
We prove this lemma along similar ideas provided in the proof of \citet[Claim 1]{OvercompleteLVMs2014}.
Let $\eta_j$'s be independent random $\pm 1$ variables with $\Pr[\eta_j = 1] = 1/2$.  We rewrite tensor $T'$ as
\begin{align} 
T' = \sum_{j>1} \eta_j a_j\otimes a_j\otimes a_j.
\end{align}
Since $a_j$'s are zero-mean random Gaussian vectors, we have $\eta_j a_j \sim a_j$, and thus, the new $T'$ has the same distribution as the original one.
We now first sample vectors $a_j$'s, and this already makes the norm $\|\cdot\|_{B*}$ well-defined. In addition, the value of $\eta_j$'s does not change the singular values of $A$ or $B$. 
Also note that since $a_j$'s are zero-mean random Gaussian vectors with expected norm 1, they also satisfy with high probability the incoherence condition such that $|\inner{a_i,a_j}| \leq \tilde{O}(1/\sqrt{d})$ for all $i\ne j$. Thus, we condition on all these fixed events, and the only remaining random variables are just the $\eta_j$'s.

The proposed statement in the lemma is equivalent to bounding 
$$\sup_{\|u\|_{B*}=1, \|v\|=\|w\|=1} \left| T'(u,v,w) \right|.$$
In order to bound it, we provide an $\epsilon$-net argument. We construct an $\epsilon$-net such that for any vector $u \in \R^d$ with unit $\|\cdot\|_{B^*}$ norm, there is a vector $u'$ in the net such that $\|B^\top(u - u')\| \le 1/k^2$. We also construct standard $\eps$-net for vectors $u,w  \in \R^d$ with unit $\ell_2$ norm. By standard construction, this $\eps$-net has size $\exp(O(d\log d))$. We now show that for all $u$ in $\epsilon$-net with unit $\|\cdot\|_{B*}$ norm, and all $v,w$ in $\epsilon$-net with unit $\ell_2$ norm, the desired bound $|T'(u,v,w)| \le \tilde{O} \left( \sqrt{k/d} \right)$ holds with high probability. Then for the other vectors $(u,v,w)$ not in the $\eps$-net, the result follows from their closest points in the net.

Now for a fixed triple $(u,v,w)$ in the $\eps$-net, we have
$$
T'(u,v,w) = \sum_{j>1} \eta_j \inner{u,a_j} \inner{v,a_j} \inner{w,a_j},
$$
which is a sum of independent random variables; recall that the randomness is from $\eta_j$'s, and $a_j$'s are already sampled and thus they are fixed here. We partition the above sum into {\em large} and {\em small} terms as $T'(u,v,w) = S_L + S_{L^c}$ such that the summation $S_L$ is the sum of large terms including terms in set
$$L := \left\{ j \in \{2,3,\dotsc,k\}: |\inner{v,a_j}| \geq \log d/\sqrt{d} \vee |\inner{w,a_j}| \geq \log d/\sqrt{d} \right\},$$
and the rest are the small terms forming $S_{L^c}$. 
Note that $|\inner{u,a_j}| \leq 1$ since $\|u\|_{B*}=1$.

{\em Bounding $|S_{L^c}|$}:
Since the variables are bounded in this summation corresponding to small terms, we use Bernstein's inequality, and thus with probability at least $1-\delta$, we have $|S_{L^c}| \leq \frac{\sqrt{k\log 1/\delta} \cdot \polylog d}{d}$ for the fixed point in the $\eps$-net. By choosing small enough $\delta = \exp(-C d \log d)$ (where $C$ is large enough constant), we can apply the union bound on the $\eps$-net, and conclude that for all the vectors in the net, $|S_{L^c}|$  is smaller than $\tilde{O}(\sqrt{k/d})$ with high probability.

{\em Bounding $|S_L|$}: Since the columns of matrix $B$ are random Gaussian vectors, it satisfies the RIP property with high probability (see Remark 3 in \citet{OvercompleteLVMs2014} for the precise definition of RIP), and thus by the definition of RIP and Lemma 3 in \citet{OvercompleteLVMs2014}, we have $\|B_L\| \le 2$ where $B_L$ is the sub-columns of matrix $B$ specified by set $L$.

We now have
$$|S_L| \leq \sum_{j \in L} |\inner{u,a_j}| \cdot |\inner{v,a_j}|\cdot |\inner{w,a_j}|
\le \sum_{j\in L} |\inner{v,a_j}| \cdot |\inner{w,a_j}| 
\le \left\|B_L^\top v\right\| \cdot \left\|B_L^\top w \right\| \le 4,$$
where the second step uses the fact that $|\inner{u,a_j}| \le 1$, the third step exploits Cauchy-Schwartz inequality, and the last step uses bound $\|B_L\| \leq 2$.
Notice that matrix $B$ is already sampled before we do the $\eps$-net argument, and therefore, we do not need to do union bound over all $u,v,w$ for this event.

Since we assume the overcomplete regime $k \geq d$, the bound on $|S_{L^c}|$ is dominant which finishes the proof.


\eprf

\end{document}